\documentclass[a4paper,10pt]{article}
 \usepackage{fullpage}

\usepackage{graphicx}
\usepackage{amsmath,amssymb} 
\usepackage{color}
\usepackage{epsfig,amsbsy}
\usepackage{algorithm,algorithmic}

\title{
\vspace{-1cm}
\begin{normalsize}
\begin{tabular}{lr}
Technical Report TR-06-2010 & \hspace{3cm} Computer Vision and Machine Learning Group\\
June 2010 & Institute of Numerical Simulation\\
& University of Bonn\\
\end{tabular}\\
\end{normalsize}
\vspace{2cm}
\textbf{Image Segmentation by Discounted Cumulative Ranking on Maximal Cliques}
}

\author{Joao Carreira, Adrian Ion and Cristian Sminchisescu\\[5mm]
Computer Vision and Machine Learning Group,\\
Institute for Numerical Simulation,\\
Faculty of Mathematics and Natural Sciences, University of Bonn\\
\{carreira, ion, cristian.sminchisescu\}@ins.uni-bonn.de}

\date{}

\begin{document}

\maketitle


\begin{abstract}
We propose a mid-level image segmentation framework that combines multiple
figure-ground hypothesis (FG) constrained at different locations and scales, into interpretations that tile the entire image. The problem is cast as optimization over sets of maximal cliques
sampled from the graph connecting non-overlapping, putative figure-ground
segment hypotheses.  Potential functions over cliques combine unary Gestalt-based figure quality scores and  pairwise compatibilities
among spatially neighboring segments, constrained by T-junctions and the boundary interface statistics resulting from projections of real 3d scenes.  Learning the model parameters is formulated as rank optimization, alternating between sampling image tilings and optimizing their potential function parameters. State of the art results are reported on both the Berkeley and the VOC2009 segmentation dataset, where a 28\% improvement was achieved.
\end{abstract}

\section{Introduction}

Segmenting an image into multiple regions has for long been considered a plausible precursor of many high level visual recognition routines. 
Indeed, if plausible image regions could be extracted so they would at least partly overlap the projections of visible surfaces in the scene, 
it would be conceivable that such interpretations can be later lifted to high-level scene percepts by invoking part-based object models and scene consistency rules.
 This has motivated research into (hierarchical) multipart image segmentations, for which many excellent methods are available \cite{Shi_2000,comaniciu_mean_shift_02,felzen04,contour_to_regions09}. But finding good multipart image segmentations in one step has proven difficult, partly due to the inherently
local nature of the grouping process. The competition constraints implicit in various methods
make it difficult to integrate scene constraints and mid-level grouping into early computations, and
can influence results in ways that do not always correlate with scene properties. Learning segmentation models has also been problematic, partly because of insufficient support for reliable feature extraction and because inference, the inner core of learning, is usually very expensive. 

The alternative computational framework we pursue assembles multipart image interpretations by tiling multiple
figure-ground image segment hypotheses using mid-level scene constraints. The problem of hypothesis selection and
consistent (full) image segmentation is formulated as optimization over sets of maximal cliques, sampled from a graph that connects non-overlapping image segments. By designing and learning clique potentials that encode both intrinsic,
unary Gestalt segment properties and pairwise spatial compatibilities that
account for plausible configurations of neighboring, spatially non-overlapping segments, we are
able to eliminate many implausible image segments and tilings that cannot possibly arise from the projection of surfaces
in typical, structured 3d scenes. We show that such a strategy achieves the state of the art in benchmarks like Berkeley and
VOC2009. 

\subsection{Related work}

Approaches to image segmentation include normalized cuts \cite{Shi_2000}, mean shift \cite{comaniciu_mean_shift_02} and minimum spanning trees \cite{felzen04}. They are usually computed multiple times, to increase the probability that some of the retrieved segments capture full objects, or their significant parts in images. Another methodology to obtain multiple segmentations is to aggregate in a hierarchy, two well-known examples being multigrid methods \cite{sharon06Hierarchy} and the Ultrametric Contour Maps \cite{contour_to_regions09}. The latter achieved state-of-the-art results in a number of challenging segmentation datasets. These algorithms partition the image into a number of regions by using pairwise pixel dependencies. Direct learning is usually targeted at finding the parameters of local affinities ~\cite{contour_to_regions09,Cour:aistats05}. Other techniques work at coarser scales by optimizing over superpixels. This allow features to be computed over a larger spatial support. Ren and Malik \cite{ren_learning_class} learn a classification model to combine superpixels based on their Gestalt properties. Hoiem et al \cite{Hoiem_2007} proposed a model that reasons jointly over scene geometry and occlusion boundaries, progressively merging superpixels so as to maximize the likelihood of a qualitative 3d scene interpretation. Instead our goal is complementary: a set of consistent full image segmentation hypotheses, computed based on mid-level Gestalt cues and implicit 3d constraints. 

While multi-part image segmentation algorithms are most commonly used, a number of figure-ground methods have been recently pursued. Bagon et al \cite{GoodSegment} proposed an algorithm that generates figure-ground segmentations by maximizing a self-similarity criterion around a user selected image point. Malisiewicz and Efros \cite{malisiewicz-bmvc07} showed that good object-level segments could be obtained by merging pairs and triplets of segments from multi-part segmentations, but at the expense of generating also a large quantity of implausible ones. Carreira and Sminchisescu \cite{carreira_cvpr10} generate a compact set of segments using parametric minimum cuts and learn to score them using region and Gestalt-based features. These algorithms were shown to be quite successful in extracting full object segments, suggesting that a promising research direction is to develop methods that combine multiple figure-ground segmentations (or just segments obtained at multiple scales, potentially from different methods), into plausible full image segmentations. Still missing is a formal multiple hypothesis computational framework for consistent selection (tiling) and learning, which we pursue here. 
{Providing a compact set of multiple hypotheses rather than a single answer is desirable for learning, for high-level, informed processing and for graceful performance degradation.}

{\paragraph{Organization:} In sec.~\ref{sec:tile} we present our maximal clique formulation framework including both the search procedure and the parameterization of the clique potentials. Sec.~\ref{sec:learning} describes our ranking-based learning framework that alternates between sampling new tilings (a discrete optimization method) and optimizing the parameters of our clique potentials (a continuous problem) against the test error measure, here the full image segmentation quality. Sec~\ref{sec:features} discusses our segment, mid-level unary and pair-wise terms based on Gestalt measures and the statistics of projected boundaries of 3d surfaces, including T-junctions and extremal edges. We show inference and learning statistics as well as experiments on the Berkeley and Pascal VOC 2009 segmentation datasets in sec.~\ref{sec:experiments}. We conclude with ideas for future work in sec.~\ref{sec:concl}.}



\section{Image tiling as sampling maximal cliques}
\label{sec:tile}

\newcommand{\s}[1]{{\mathcal #1}} 

\newcommand{\Q}{\f{Q}}
\newcommand{\G}{\s{G}}
\newcommand{\V}{\s{V}}
\newcommand{\E}{\s{E}}
\newcommand{\U}{\s{U}}

\newcommand{\w}{ \boldsymbol{\theta} }
\newcommand{\wu}{\w_u}
\renewcommand{\wp}{\w_p}
\newcommand{\ws}{\theta} 

\newcommand{\xu}{\boldsymbol{\phi}_u} 
\newcommand{\xp}{\boldsymbol{\phi}_p}

\newcommand{\f}[1]{{\mit #1}}

\renewcommand{\sf}{\f{F}_\ws} 
\newcommand{\of}{\f{S}_\ws} 

\newcommand{\T}{c} 
\renewcommand{\P}{\s{C}} 

\renewcommand{\S}{\s{H}} 

\newcommand{\fgtiling}{\textit{FG-Tiling}}
\newcommand{\randtiling}{\textit{Constrained-random}}
\newcommand{\enumonemin}{\textit{Enum-1min}}

Given a set $\S$ of $N$ segments  our aim is to generate several tilings $\T$ such that no two segments on $\T$ overlap and $\T$ has a high score $\sf(\T)$.
Consider for that a graph $\G(\V, \E)$, called the \textit{consistency graph}, where the vertices are the segments in $\S$. Two vertices are connected by an edge if the corresponding segments do not overlap.\footnote{While disallowing overlap increases the exposure to imperfect
boundary alignments between the available segments, it leads to a dramatic reduction in the solution
space and doesn't require additional processing to assign pixels lying on the intersection
of overlapping segments.} A \textit{clique} of $\G$, which is a fully connected subgraph of $\G$, corresponds to a set of segments that can form a tiling. A clique is called \textit{maximal}\footnote{Also called inclusion maximal clique.} if it is not included into any other clique and hence a larger clique cannot be obtained by adding vertices to it. In our case a maximal clique corresponds to a tiling that cannot be extended using any other segment in $\S$. A \textit{maximum clique} of a graph is a clique with the largest number of vertices. A \textit{maximum weighted clique} is a clique that maximizes the sum of weights associated to its vertices.

We formulate the search for tilings as finding \textit{maximal cliques} $\T \subseteq \G$ with high potential $\sf(\T)$
\begin{equation}\label{eq:score-tiling}
\sf(\T) = \sum_{s_i \in \T} \wu^T \cdot \xu(s_i) + \sum_{s_i \in \T, s_j \in \T \cap \mathcal{N}(s_i)} \wp^T \cdot \xp(s_i, s_j)
\end{equation}
where $\xu(s_i)$, $\xp(s_i, s_j)$ are feature vectors extracted for, respectively, segment $s_i$ and \textit{image neighbors} $s_i, s_j$ (denoted $s_j \in \mathcal{N}(s_i)$). $\w = (\wu, \wp)$ are the corresponding weights learned as mentioned in Section~\ref{sec:learning}.

The problem of finding the \textit{maximum} (weighted) clique of a general graph is known to be both NP-complete and hard to approximate to a given bound~\cite{Bomze99themaximum}. Existing algorithms produce one single solution which equals or approximates the maximum clique. In the weighted case maximization is done only over unary terms associated to vertices. This is different from our case. We desire multiple tilings for each image and the potential of a clique (tiling) depends on both unary and pairwise terms. Enumerating all cliques to find the optimum is not feasible as we deal with many vertices (over 150) and the complexity to enumerate all cliques of size $k$ of a graph with $N$ vertices is $O(N^k k^2)$. Finding a \textit{maximal} clique can be done in linear time in the number of vertices, by starting with one vertex and adding each of the other vertices in some order. But graphs that have a large maximum clique can have maximal cliques of arbitrary small size. To obtain multiple estimates we follow a two step greedy approach: {\emph(i)} starting with each vertex generate a maximal clique; {\emph(ii)} refine each solution using a local search in the space of maximal cliques based on the trained cost function. We generate up to $N$ different tilings, ranked in decreasing order of $\sf$. Notice that our approach is based on established strategies to find approximations of the maximum clique (step 1 is known as a \textit{sequential greedy heuristic} and step 2 as a \textit{local search
heuristic}~\cite{Bomze99themaximum}). \textit{Algorithm}~\ref{alg:greedy-tiling} describes the proposed method.

\begin{algorithm}[tbp]
\caption{\fgtiling($\S_I, \w, \xu, \xp$) - Discrete optimization for image tilings.}\label{alg:greedy-tiling}
\textbf{Input}: Pool of $N$ segments $\S$, weights $\w = (\wu, \wp)$, features $\xu, \xp$.\\[-\baselineskip]
\begin{algorithmic}[1]
\STATE $\{s_i\}_{i=1\dots N} \leftarrow$ segments in $\S$ in decreasing order of $\wu^T \cdot \xu(s_i)$ \COMMENT{unary terms}\label{lin:start}
\FOR {$i = 1 \dots N$}
  \STATE $\T_i \leftarrow \{s_i\}$ \COMMENT{initialize clique}
  \STATE \COMMENT {Step 1: sequential greedy heuristic to ``build'' maximal clique}
  \FOR {$j = 1 \dots n$} \label{lin:step1-st}
  \IF {$s_j$ does not overlap any segment in $\T_i$} \label{lin:step1-test}
    \STATE $\T_i \leftarrow \T_i \cup \{s_j\}$
  \ENDIF
  \ENDFOR \label{lin:step1-end}
  \STATE \COMMENT {Step 2: local search heuristic for solution refinement}
  \REPEAT
  \FORALL {$s' \notin \T_i, s'$ not overlapping $s_i$}

  \STATE $\T' \leftarrow (\T_i \setminus \mathcal{O}(s')) \cup \{s'\}$ \COMMENT{remove segments that overlap $s'$, add $s'$}

  \STATE $\T' \leftarrow \T' \cup \{s_{l1}, s_{l2}, \dots\}$ \COMMENT{extend $\T'$ to a maximal clique like in lines~\ref{lin:step1-st}--\ref{lin:step1-end}}
  \IF {$\sf(\T') > \sf(\T_i)$ \COMMENT{see eq.~\ref{eq:score-tiling}}}
  \STATE $\T_i \leftarrow \T'$
  \ENDIF
  \ENDFOR
  \UNTIL {convergence}
\ENDFOR
\end{algorithmic}
\textbf{Output}: Pool of tilings $\P = \cup \{\T_i\}$ for the current image ranked in decreasing order of $\sf(\T_i)$.
\end{algorithm}


\paragraph{Complexity:} The size of the largest clique that can be formed with a certain vertex is bounded by the degree of this vertex, in our case $d = \deg(s_i)$.
If a set $\U$ is kept, containing the segments in $\S \setminus \T_i$ which are not overlapping any segment in $\T_i$, the complexity of {step 1} is $O(N + d^2)$. 
Maximum $N$ steps are needed to build $\U$ from the list of sorted segments and $d^2$ is an upper bound for the loop in step 1 and the verification inside.

{Step 2} can be executed in $O(M d (d + N+d^2))$ where $M$ is the maximum number of iterations allowed\footnote{In experiments we use $M=10$.}. The inner loop over all $s'$ is bounded by $d$ as $s'$ must not overlap $s_i$. Rejecting segments in $c'$ overlapping with $s'$ is also bounded by $d$ as all segments previously in $c'$ are not overlapping $s_i$. Finally, extending $c'$ to a maximal clique has the same complexity as step 1, namely $O(N + d^2)$.

Ordering the segments is done only once, thus the complexity for running \fgtiling \ for all segments $s_i \in \S$ is $O(N \log N + N (N + d^2 + M d (d + N + d^2)))$ where the dominant {worst case component} is $O(N d^3)$ if $M$ is fixed. In practice our matlab implementation using $M = 10$ takes on average 20 seconds per image for the BSDS test set.

%
%
%
%

\section{Learning mid-level vision}
\label{sec:learning}

Assume we are given a set of features $\xu(s_i), \xp(s_i, s_j)$ computed, respectively, for segments $s_i \in \S$ and pairs of segments $s_i, s_j \in \S$ which are neighbors in the image, i.e. $s_i, s_j$ share a common boundary and do not overlap. We search for the weights $\w = (\wu, \wp)$  such that the ranking of tilings induced by $\sf$ (eq.~\ref{eq:score-tiling}) is as close as possible to the ranking induced by the quality of the tilings with respect to the ground truth.

\begin{algorithm}[tbp]
\caption{Learning algorithm that estimates parameters $\w = (\wu, \wp)$.}\label{alg:learning}
\textbf{Input}: Segments $\S_I$ for the images in the training set $T$, features $\xu, \xp$, rank $K$.\\[-\baselineskip]
\begin{algorithmic}[1]
\STATE $\w \leftarrow (\w_{u0}, \w_{p0})$ \COMMENT {Initialize weights}
\FORALL {$I \in T$}
  \STATE $\P_I \leftarrow$ \fgtiling($\S_I, \w$) \COMMENT{extract initial tilings for image $I \in T$}
\ENDFOR
\REPEAT \label{lin:outer-start}
  \STATE $\w \leftarrow \arg\max_\theta\{\sum_{I \in T} \of(\P_I, K)\}$ \COMMENT{optimization: find $\w$ that maximizes $\of$} \label{lin:inner-loop}
  \FORALL {$I \in T$}
    \STATE $\P_I \leftarrow$ \fgtiling($\S_I, \w, \xu, \xp$) \COMMENT{extract new tilings for image $I \in T$}
  \ENDFOR
\UNTIL {no improvement} \label{lin:outer-end}
\end{algorithmic}
\textbf{Output}: Weights $\w = (\wu, \wp)$.
\end{algorithm}

The learning process alternates between the discrete optimization of tilings, where it runs \fgtiling \ with the existing parameters $\w$ to create a new pool of tilings for each of the images in the training set, and a continuous parameter optimization step that finds parameters $\w$ which maximize an objective function $\of$ on the produced tilings, as used for testing: the overlap with ground truth (\textit{Algorithm}~\ref{alg:learning}). Instead of aiming to enforce only the best tiling in the first position, which might be impossible, we design a scoring (with best as special case) that aims at ranking tilings in decreasing order of their quality.
For an image $I$, weights $\w$, and a pool of tilings $\P_I = \{\T_1,\T_2, \dots\}$ where $\T_i$ is the tiling at rank $i$ when sorting $\P_I$ in decreasing order of the value of $\sf(\T_i)$, the objective function $\of(\P_I)$ is:
\begin{equation}\label{eq:weighting}
\of(\P_I, K) = \sum_{i=1}^K w(i)Q(\T_i)
\end{equation}
where $Q(\T_i)$ is the quality of $\T_i$ measured using the ground truth, $w(i)$ is the weighting of rank $i$, and $K$ is the rank parameter which determines the constraint we want to enforce (e.g. $K = 1$ for only the best ranked, $K = |\P_I|$ for a full K-ordering). We define $Q(\T_i)$ as the average \textit{covering} of $\T_i$ with all ground truth segmentations as in ~\cite{contour_to_regions09}. The covering is the sum of overlaps between each individual segment in a ground truth segmentation and the closest segment in a tiling, multiplied by the area of the ground truth segment. $O(s_i, s_g) = |s_i \cap s_g| / |s_i \cup s_g|$ is the standard overlap measure between $s_i$ and $s_g$~\cite{pascal-voc-2009}. For rank weighting, we use: $w(i) = w(K,i) = 1/[1+(i-1)/(K-1)]$. This decay is similar to the \textit{Discounted Cumulative Gain} (DCG)~\cite{JarvelinK02} that uses a logarithmic reduction factor of the form $1/\log_2(i+1)$. DCG penalizes more aggressively the error in the first ranks. We found this to work slightly less well in our tests.\footnote{For an image segment graph $\G$ with $N$ nodes, clique potentials can be used to define a constrained probability distribution over partitions.  We can write a Gibbs distribution over cliques as $p_{\w}(\T)= \exp{(\sf(\T))}/\sum_{\T \in \G} {\exp(\sf(\T))}$, and can learn using ML, with partition functions approximated by summing only over $O(N)$ cliques, computed by \fgtiling \ (This approach will be presented in an upcoming technical report.). Here we choose a different loss that directly optimizes the overlap measure used during test time. Notice however, our very different use of cliques compared to product expansions in graphical models. Along this path, modeling the nodes as binary variables in a random field would neither produce the semantics we need, nor would necessarily lead to clique consistent inference.}


\section{Mid-level image descriptors}
\label{sec:features}
Our model aims to generate full image tilings that have properties similar to the ones of ground truth segmentations produced by human annotators. We use both unary features inspired by Gestalt properties and pairwise features sensitive to the boundary statistics arising from projections of 3d surfaces, for a total of $46$ unary and $22$ pairwise features. These features are computed once and do not change during learning and inference. All features are individually normalized to zero mean and standard deviation $1$.

\noindent{\bf Unary Descriptors:} As unary features, we primarily use the ones proposed in \cite{carreira_cvpr10}, that include the amount of \textbf{contrast along the boundary} of the segment (8 features), $\xu^b(s_i)$, region properties such as position in the image, area and orientation, $\xu^r(s_i)$ (18 features), as well as \textbf{Gestalt} properties such as convexity and dissimilarity between the segment interior and the rest of the image in terms of intensity and texture, $\xu^d(s_i)$ (8 features).

We complemented the unary features in \cite{carreira_cvpr10} with a novel set of $12$ responses quantifying \textbf{center-surround dissimilarity}, $\xu^l(s_i)$. We define three image strips of width $18$, $30$ and $42$ pixels around each segment. We compute how dissimilar each strip and the segment are according to $4$ different local features: hue, rgb, SIFT and textons. For each type of local feature and each strip, dissimilarity is determined as the chi-square distance between the histogram of quantized local features in the strip and in the segment, resulting in the $12$ features. The local features are sampled on a regular grid, every $10$ pixels. The color histograms use patches $4$ and $8$ pixels wide, while the SIFT patches are $8$ and $18$ pixels wide. The textons are the ones used in globalPb ~\cite{contour_to_regions09} quantized into $64$ bins. We quantize the other features into $30$ bins, with the codebook being obtained \textit{in each image} at test time by k-means. \\

\noindent{\bf Pairwise Descriptors:} We define a segment neighborhood between pairs of segments sharing a boundary and not overlapping. The occurrence of such pairs is usually non-accidental, particularly in our pool of figure-ground segmentations, because we don't consider the \textit{ground}. Segments that are artifacts of the particular parameter and location constraint that generated them will tend to have few neighbors. Computing this type of neighborhoods can be done robustly by growing all segments by a small amount (4 pixels in our implementation) and then detecting the pairs that overlap.
The pairwise features capture the configuration of pairs of segments. We use two sets of pairwise features. The first encodes \textbf{pairwise region properties} such as relative area, position and orientation and is simply defined by $\xp^r(s_i,s_j) = |\xu^r(s_i)-\xu^r(s_j)|$ (18 features).

\begin{figure}[tbp]
\label{fig:tjunction}
\centering
\mbox{
\epsfig{figure=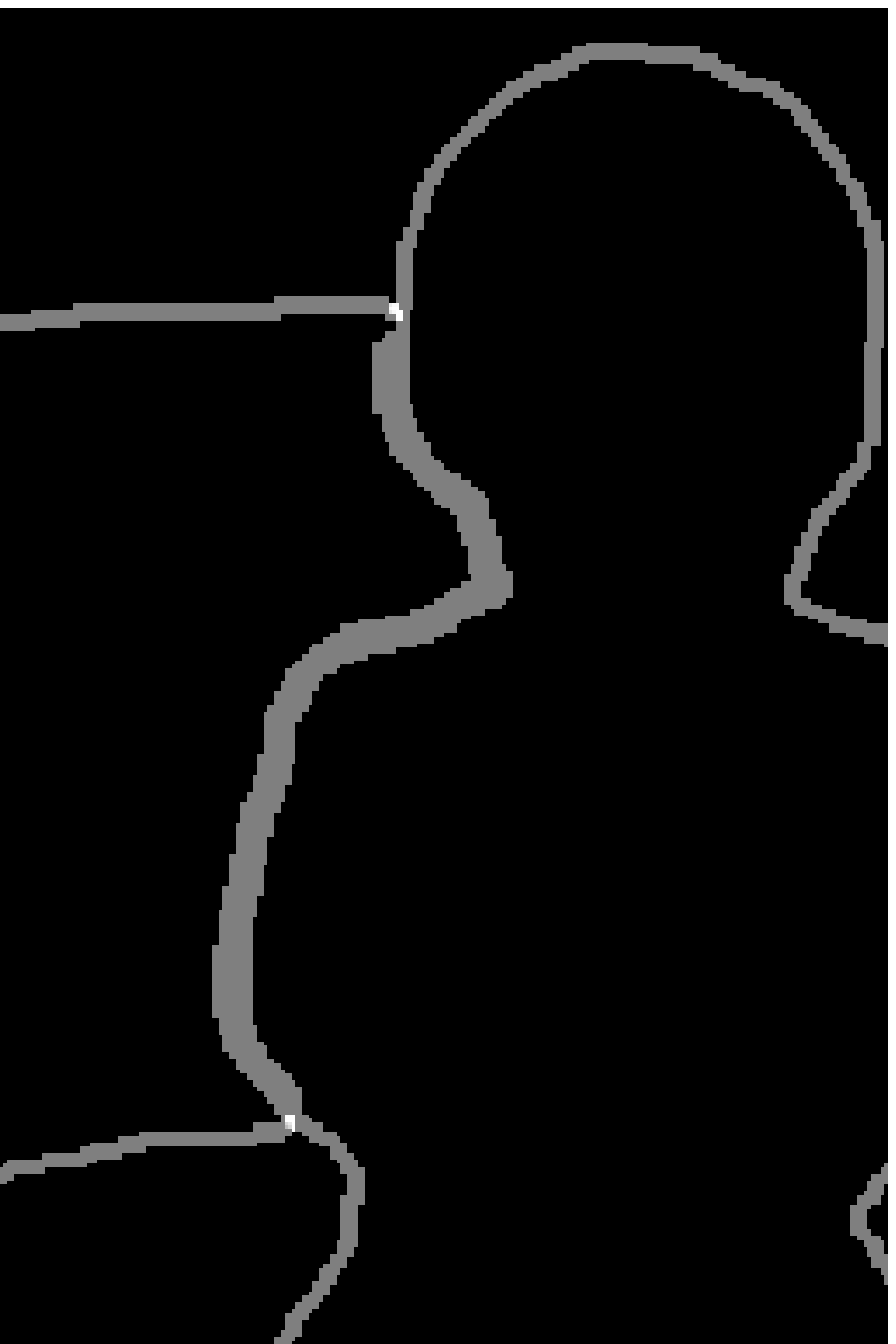,height=0.18\textwidth}~
\epsfig{figure=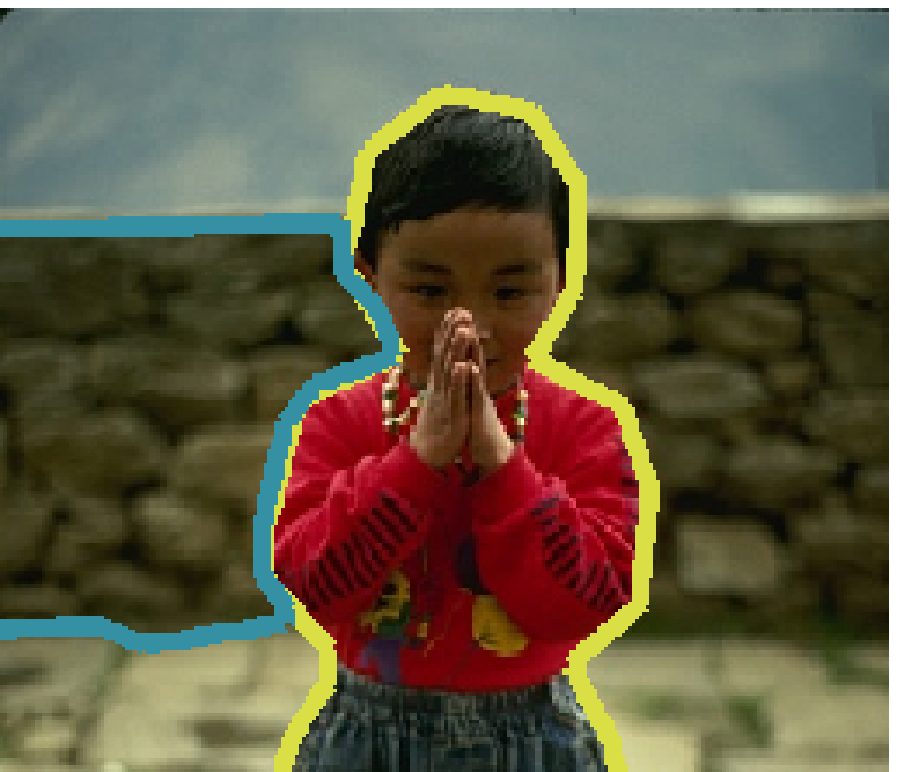,height=0.18\textwidth}~
\epsfig{figure=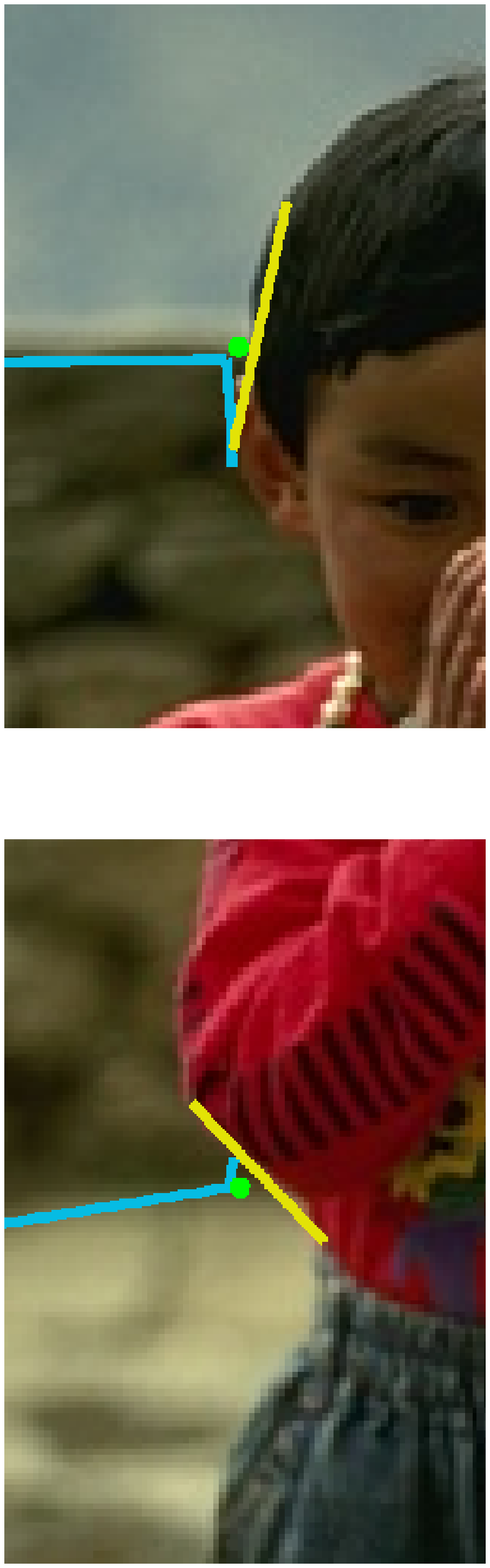,height=0.18\textwidth}~
\epsfig{figure=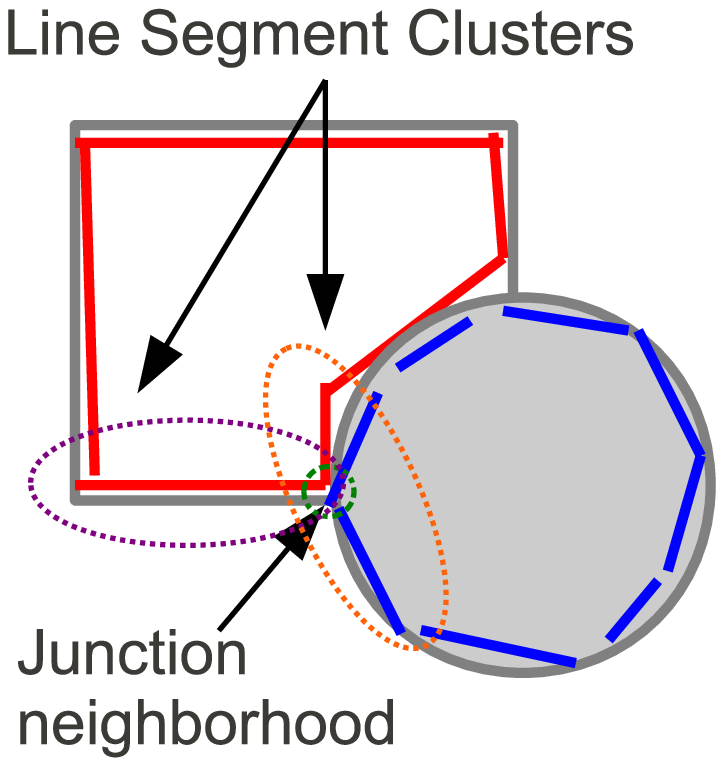,height=0.18\textwidth}~
\epsfig{figure=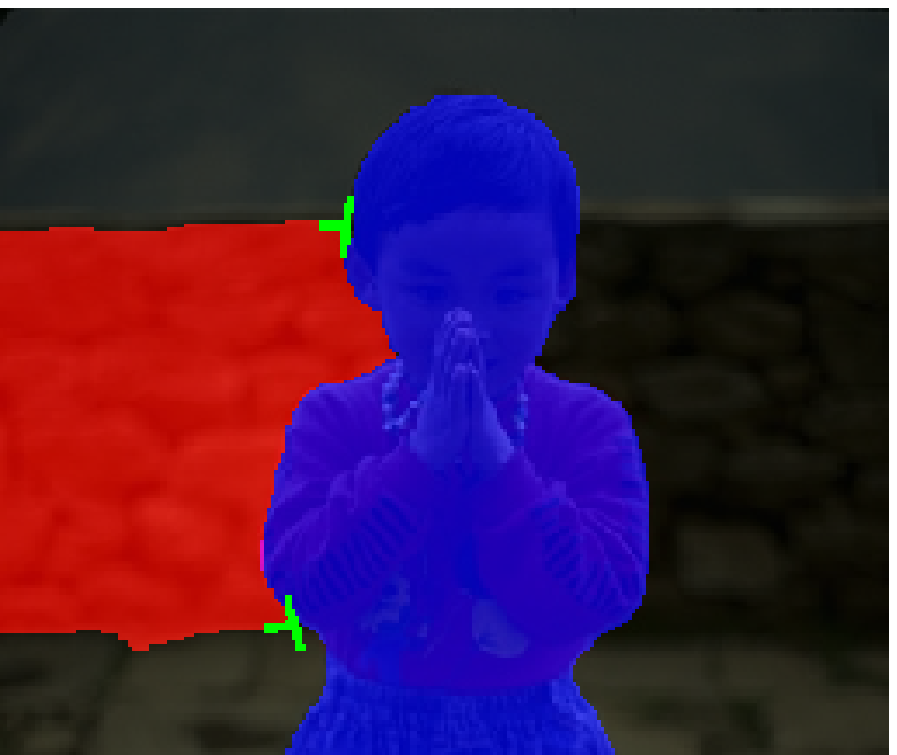,height=0.18\textwidth}~
}
\caption{Our T-junction detector works on all pairs of non-overlapping and spatially neighboring segments. In order to detect junctions, we grow the two regions plus their shared background, sum the three binary masks, and find the points in the image where the sum is maximized (first image on the left). 
 These are initial junction points, and are improved by solving a least squares problem minimizing the distance to the closest line segments approximating the boundaries of the two regions (image $2$ and $3$). To form the base and the leg of the T, these line segments are clustered into two sets based on their orientation, using agglomerative clustering, and a line is fit to each cluster (image $4$). The cluster having a line segment endpoint with maximum minimal deviation from the junction along its fitted line is set as the base of the T. The final result is shown on the last image on the right.}
\end{figure}

We also employ $4$ features which signal occlusion. In ground truth segmentations, neighboring segments often correspond to projections of objects at different depths, which result in distinctive image statistics. These are sufficiently informative even for determining which of the two neighboring regions corresponds to the occluding surface in 3D space, the so called figure-ground assignment problem 
\cite{RenFM06} \cite{Leichter_09}.
The occluding segment usually has a higher convexity coefficient and is often surrounded by the occluded segment. Let $a(s_i)$ be the unary convexity feature in $\xu^d(s_i)$. Then the \textbf{relative convexity} feature  is implemented as $\xp^a = |a(s_i) - a(s_j)|$. Let the length of the adjacent boundary between two segments be $l_{12}$, and the segment perimeters be $l_1$ and $l_2$. Then \textbf{surroundedness} is defined as $\xp^s = |l_1/l_{12} - l_2/l_{12}|$.
Another important occlusion features are \textbf{t-junctions}, boundary patterns shaped as a T, usually caused by the intersection of the boundaries of two objects in an occlusion relationship. Typically the location of the leg of the T indicates which segment is occluding the other. T-junctions were used in recent approaches to figure-ground assignment, as an energy term for triplets of regions in CRFs \cite{RenFM06} \cite{Leichter_09} \cite{Hoiem_2007_5826}. Here we model them directly as a pairwise segment compatibility feature, by measuring the consistency with which the leg of the t-junctions belongs to the same segment, weighted by the quality of the fitting of the junction to a T, as opposed to being Y-like. The feature is defined as $|\xp^t = \sum_k{[b_i(t_k) - b_j(t_k)]}|$, with the sums being over all junctions between the pair of segments. The weighting is $b_i(t) = \exp(-|(\pi/2 - \alpha_t|)$, $\alpha_t$ being the angle formed by the leg of the junction $t$ with the base. When the leg of the junction is on the boundary separating both segments, or the leg is not on the boundary of segment $i$ then $b_i(t)$ is set to $0$. 
Junctions are hard to detect when considering pixel intensities locally, even for humans \cite{McDermott2002}. But given a pair of neighboring segments this can be done robustly, as illustrated in fig.~\ref{fig:tjunction}.

The \textbf{shading along region borders} was shown to provide information about occlusion in both computational \cite{Huggins01} and
psychophysical tests, under the name of \textit{extremal edges} \cite{GhoseTandra2005}. The phenomenon is explained by the illumination gradient tending to be orthogonal to the boundary, on the occluding side. We implement the gradient orthogonality feature $o(s_i, s_j)$ as in \cite{Leichter_09} and produce the compatibility  feature as $\xp^e(s_i,s_j) = |o(s_i, s_j) - o(s_j, s_i)|$. The absolute value is computed because we're not interested here in determining which segment is in front, just in having an occlusion indicator. 

\section{Experiments}
\label{sec:experiments}

Our inference and learning methods were tested on the Berkeley Dataset (BSDS) \cite{MartinFTM01} and on the Pascal VOC 2009 Segmentation Dataset (VOC2009) \cite{pascal-voc-2009}.
For comparison we show results of the Oriented Watershed Transform Ultrametric Contour Maps using globalPb as contour detector (gPb-owt-ucm) \cite{contour_to_regions09}.

We generate a pool of segments using the publicly available implementation of Constrained Parametric Min-Cuts (CPMC) \cite{carreira_cvpr10},
which produces nested sets of segments around rectangular seeds on a regular grid with predicted qualities for each segment.
Per image an average of 194 segments is generated for the BSDS test set and 156 segments for the VOC2009 validation set.
This algorithm was recently shown to produce compact sets of segments that accurately cover ground truth objects. 

\begin{figure}[tbp]
\centering
\includegraphics[width=4.8cm]{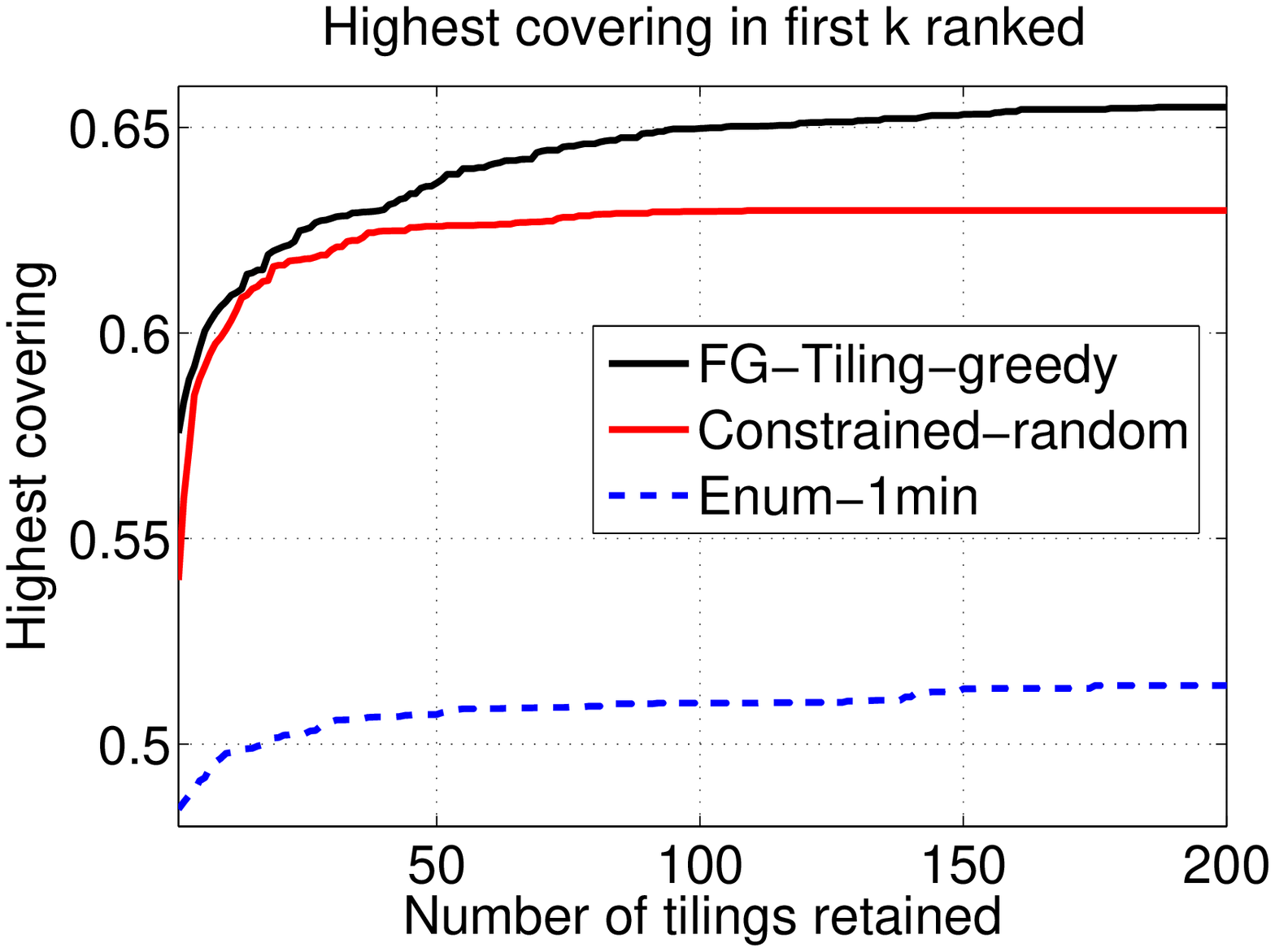}
\hspace{-4mm}
\includegraphics[width=4.8cm]{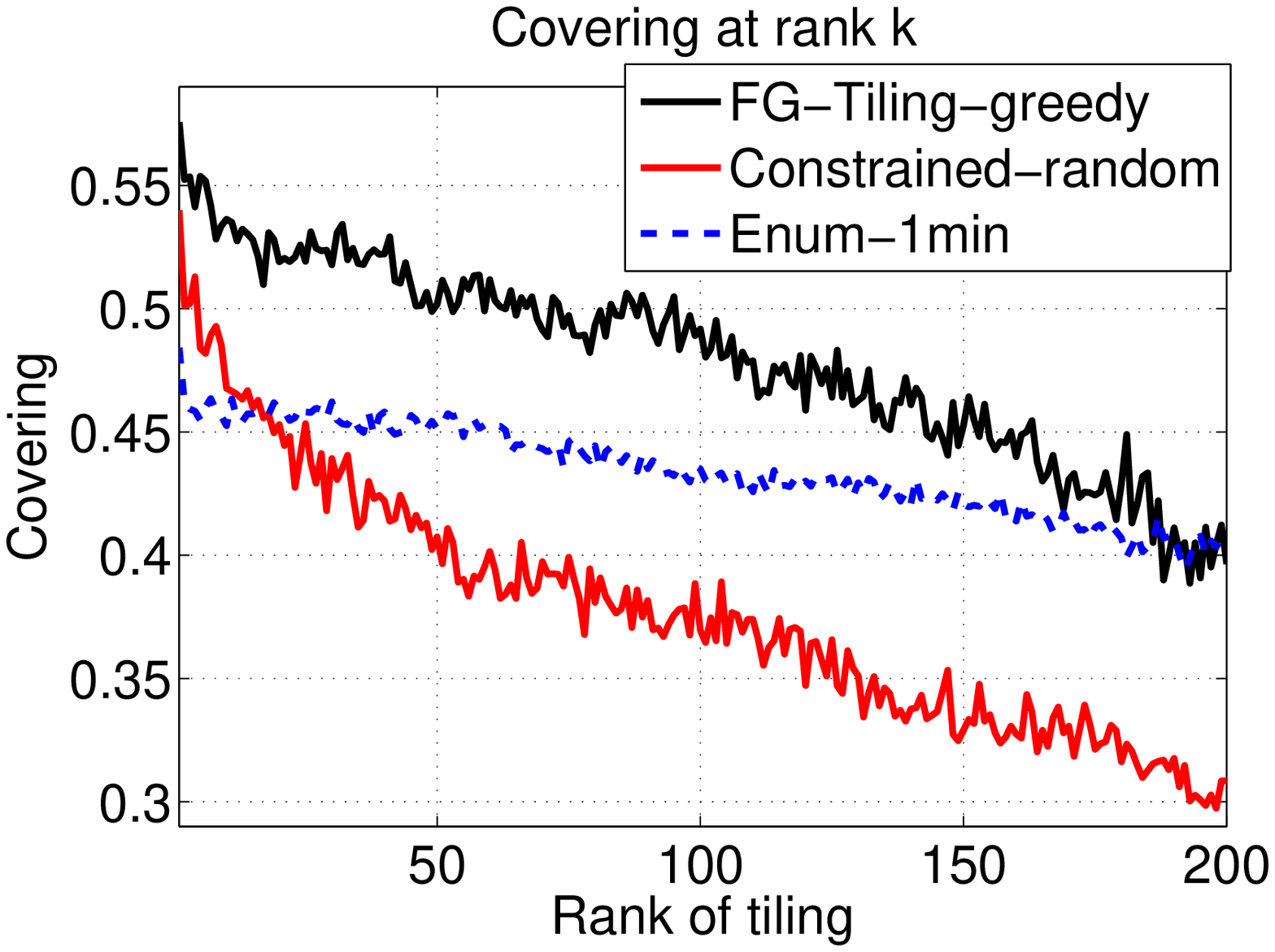}
\hspace{-4mm}
\includegraphics[width=4.8cm]{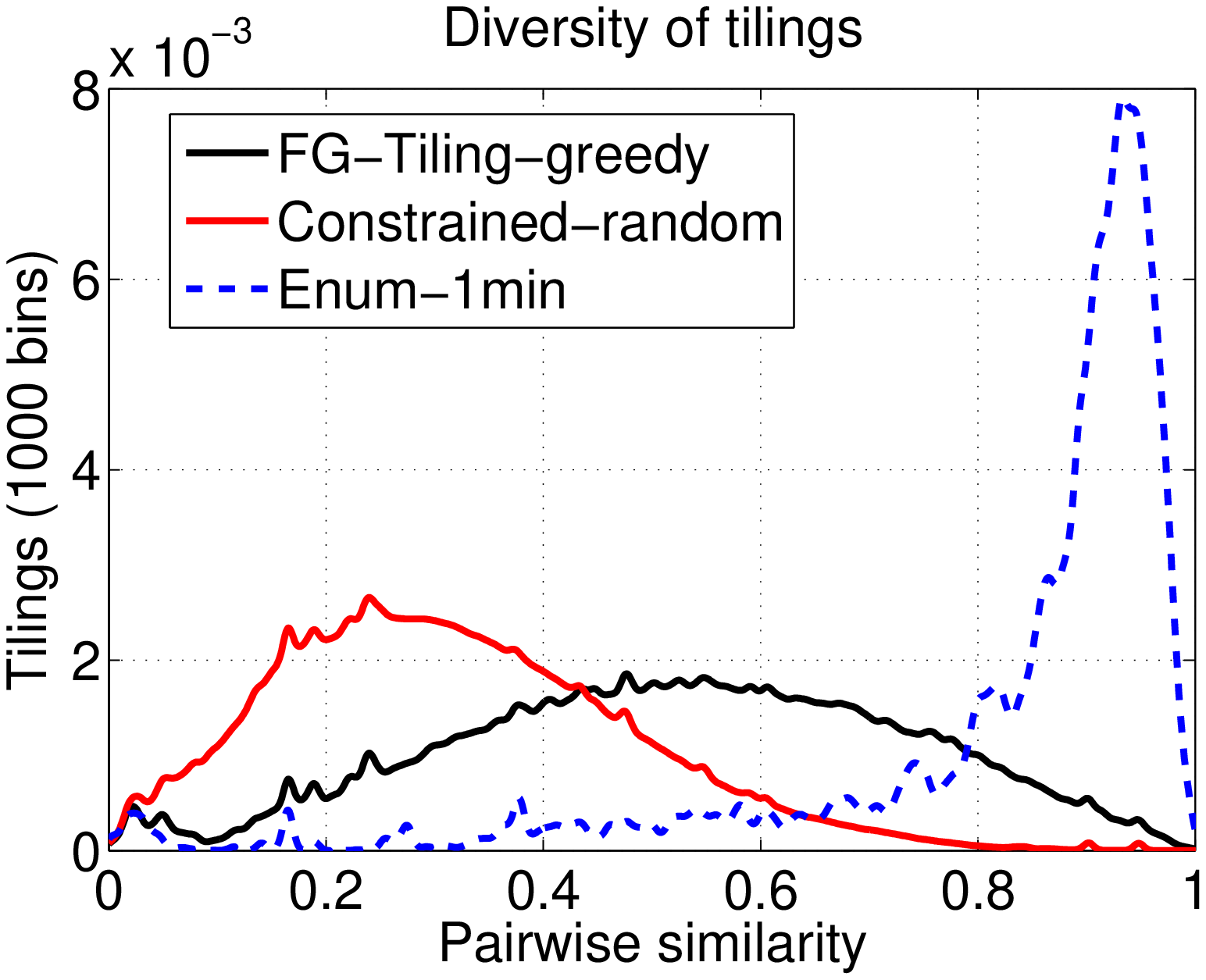}\\
\vspace{-2mm}
\caption{Evaluation of \fgtiling, \randtiling, and \enumonemin \ on the BSDS dataset. \textit{Left}: highest quality $Q(\T)$ for number of tilings considered. \textit{Center}: average quality for given rank (if exists). The quality $Q(\T)$ was measured with respect to the ground truth using the \textit{covering} measure (see sec.~\ref{sec:experiments} for details). \textit{Right}: histogram of pairwise similarity between produced segmentations. 
}
\label{fig:plots-tilings}
\end{figure}

Fig.~\ref{fig:plots-tilings} shows the evaluation of \fgtiling \ and two baselines, \enumonemin \ and \randtiling, on the BSDS dataset (see sec.~\ref{sec:experiments}). All methods produce maximal cliques i.e. tilings with segments that do not overlap and the cliques cannot be extended using the current pool of segments. For each method the produced tilings are ranked using the scoring function in eq.~\ref{eq:score-tiling}. 

\enumonemin \ is an algorithm that recursively, exhaustively, enumerates maximal cliques until the given time of 1 minute per image is reached and returns the highest scoring $N$ cliques that have been found\footnote{The time of 1 minute given to \enumonemin \ is equal to 3 $\times$ the average time of \fgtiling \ on the BSDS test set. Without the time constraint the algorithm did not finish enumerating cliques after 48 hours on a test image where a pool of $N=120$ figure-ground segmentations had been used.}. Similar to line~\ref{lin:start} of \fgtiling, \enumonemin \ first sorts the segments based on $\wu^T \cdot \xu(s_i)$. During enumeration, it quickly finds one tiling similar to the result of step 1 in \fgtiling. However, within 1 minute, it produces only small variations of the same tiling, as seen also in fig.~\ref{fig:plots-tilings}, right. \randtiling \ is similar to step 1 in \textit{Algorithm}~\ref{alg:greedy-tiling} with the difference that in line~\ref{lin:start} the order of the segments is randomized. The method gets a few ``lucky shots`` which explains the quite high values in the plot in fig.~\ref{fig:plots-tilings} left, but overall the average quality of the produced tilings is much lower than the other two methods (23\% less than \fgtiling \ on the test set of BSDS). \fgtiling \ balances the diversity and quality of the produced tilings to give the best results of all methods.

During learning, for the initial run of \fgtiling \ we set the weights $\wp$ corresponding to the pairwise terms to zero. The weights $\wu$ corresponding to the unary terms are set using linear regression s.t. $\wu^T \cdot \xu(s_i)$ approximates the response $\max_{s_g \in G_I} O(s_i, s_g)$ where $G_I$ is the set of ground truth segments for the image. Parameter optimization is done using a Quasi-Newton method.
During this step, the sum of $\of$ over all images in the training set and their corresponding pools of tilings is maximized. The first time this step is executed, the initial weight estimates $\w = (\wu, \wp)$ required to initialize the search are obtained using linear regression over all tilings produced for the training set. Regression uses targets $Q(\T)$ for each tiling $\T$.
The inner loop (line~\ref{lin:inner-loop}) needs on average 15 iterations to converge. The outer loop (lines~\ref{lin:outer-start}--\ref{lin:outer-end}) saturates after a few iterations (3--4) and both the quality of the first ranked tiling as well as the highest quality over all tilings for each image are maximized.
\begin{figure}[tbp]
\centering
\includegraphics[width=4.8cm]{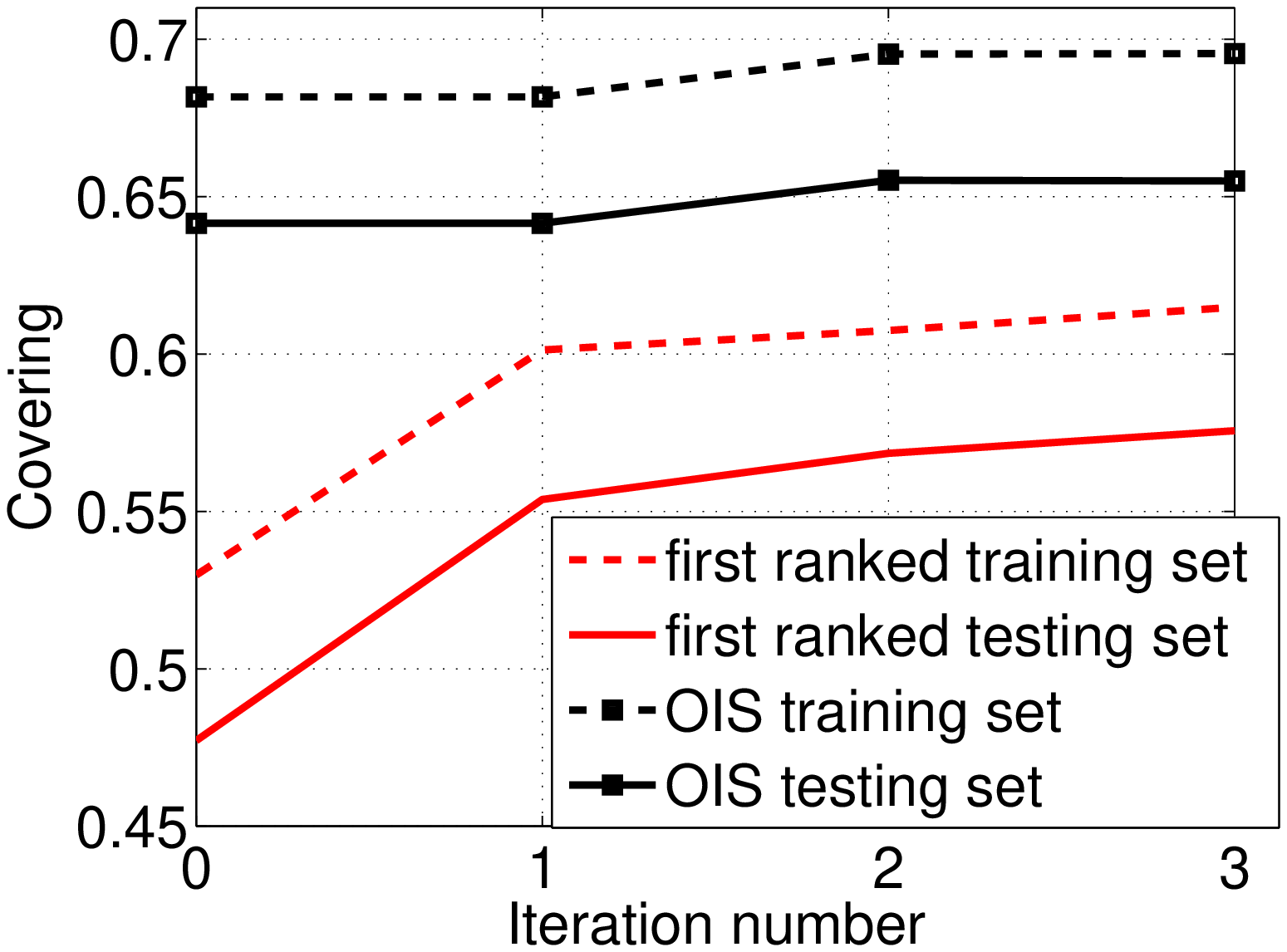}
\hspace{-4mm}
\includegraphics[width=4.8cm]{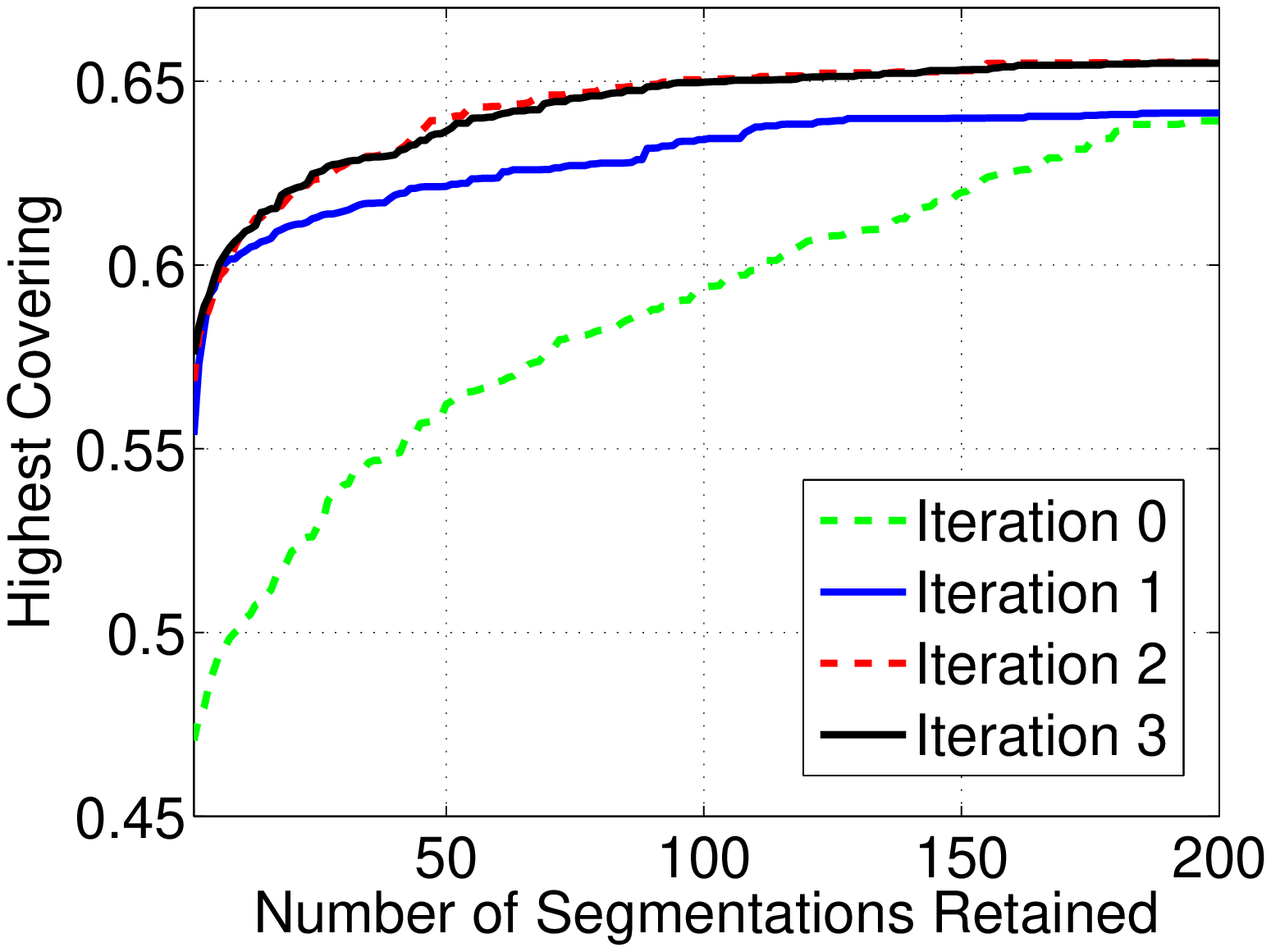}
\hspace{-4mm}
\includegraphics[width=4.8cm]{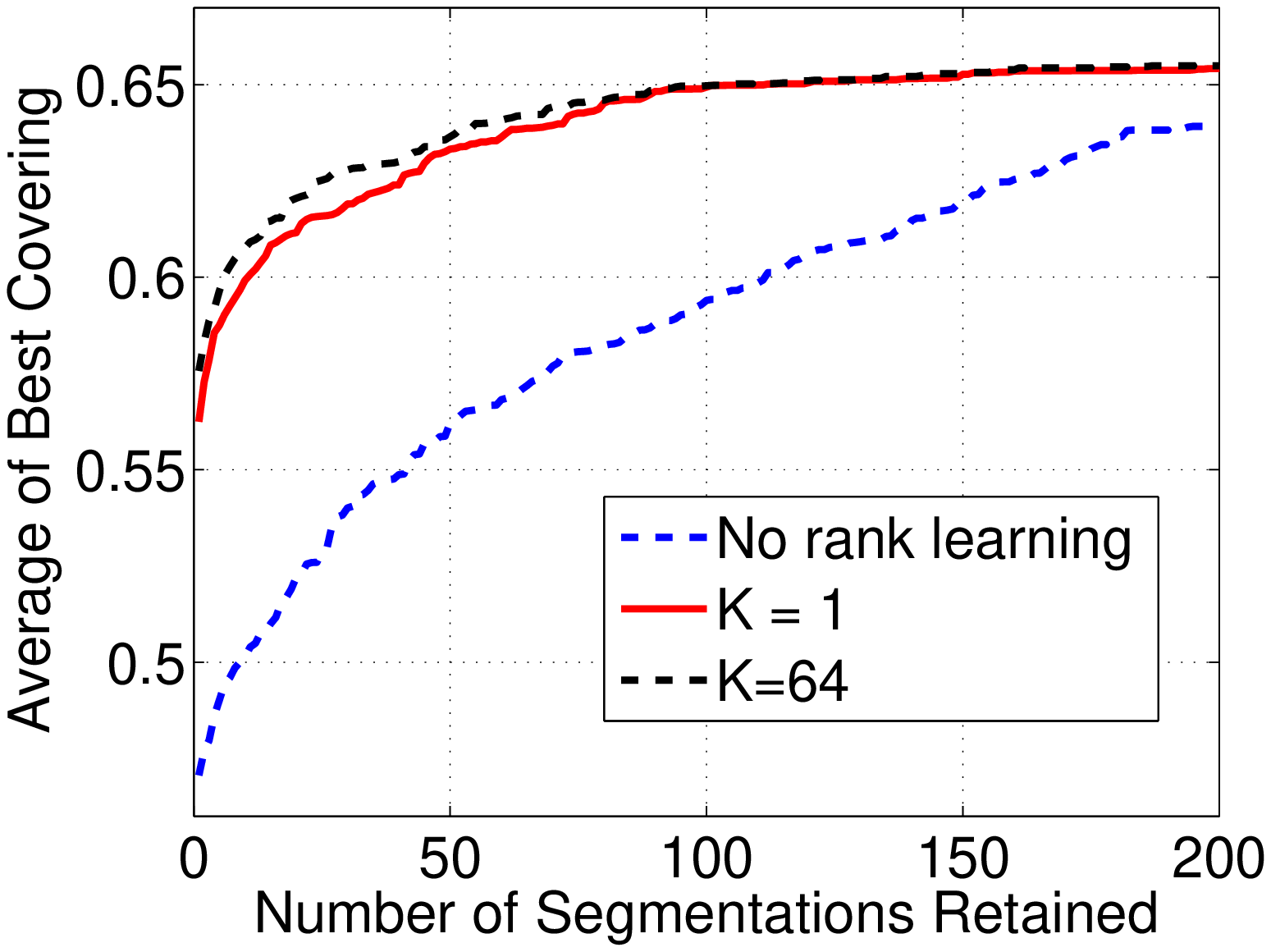}\\
\vspace{-2mm}
\caption{\textit{Left, center}: progress of learning for $K=64$ rank optimization on the BSDS dataset. \textit{Left}: progress of the first ranked and the highest quality tilings on the training and testing sets. Iteration 0 corresponds to the results with the initial weights $\w_{u0}, \w_{p0}$, iteration 1: the same tilings after the first optimization step, iterations 2--3: after new tilings and learned weights. \textit{Center}: highest quality vs. number of segmentations retained on the BSDS test set. \textit{Right}: highest quality vs. number of segmentations retained with no rank learning, learning with rank parameter $K=1$ and $K=64$.}
\label{fig:plots-learning}
\end{figure}
Fig.~\ref{fig:plots-learning} shows the progress of learning on the Berkeley Segmentation Dataset (BSDS)~\cite{MartinFTM01} using $K=64$ and a comparison of the results: without learning, learning with $K=1$ and with $K=64$. We observe that compared to $K=1$, $K=64$ produces a slightly better ranking also on the first position, presumably due to the additional constraints from lower ranks.

\begin{table}[tbp]
\centering
\begin{tabular} {c|c|c|c|c}
\hline\hline
BSDS & OIS & First & ODS & BIS \\ [0.5ex]
\hline
max. possible & 0.73 & 0.73 & 0.73 & 1.00\\
\hline
gPb-owt-ucm & 0.64 &- & 0.58  & 0.74 \\ [1ex]
\fgtiling{} & 0.64 & 0.58 & -  & 0.78 \\
\hline
\end{tabular}
\hspace{5mm}
\begin{tabular} {c|c|c|c|c}
\hline\hline
VOC2009 & OIS & First & ODS &  BIS \\ [0.5ex]
\hline
max. possible & 1.00 & 1.00 & 1.00 & 1.00\\
\hline
gPb-owt-ucm & 0.58 & - & 0.45 & 0.61 \\ [1ex]
\fgtiling{} & 0.74 & 0.52 & -  & 0.78 \\
\hline
\end{tabular}
\caption{Average coverings on the test set of BSDS and on the validation set of VOC2009. The OIS scores for \fgtiling{} are obtained considering a maximum of 64 respectively 73 tilings per image, which equals the average number of segmentations produced by gPb-owt-ucm~\cite{contour_to_regions09} on the BSDS test respectively VOC2009 validation sets (notice however that our method uses considerably fewer segments, on average 194 respectively 156 as opposed to 1100 and 1043 in gPb-owt-ucm). If \fgtiling{} uses the same number of segments but more tilings (on average 176 and 140 in BSDS and VOC2009 respectively), OIS scores of 0.66 respectively 0.76 are obtained.}
\label{table:results_voc_berkeley}
\end{table}

Table~\ref{table:results_voc_berkeley} shows results of benchmarks on the test set of BSDS and on the validation set of VOC2009.
The values represent average covering scores of ground truth segmentations by the output segmentations.
BIS measures the best covering of the ground truth segmentations by individual segments from any segmentation produced by the evaluated method.
OIS and ODS have been used in~\cite{contour_to_regions09} to evaluate the results of gPb-owt-ucm.
They have been introduced in the context of hierarchical segmentation, where scale is used to navigate from coarser to finer segmentations.
The optimal image scale (OIS) measures for each image the quality of the produced segmentation that best covers the ground truth.
The optimal dataset scale (ODS) measures the quality of the segmentations when the same scale is selected for all images. The scale to be evaluated is chosen to maximize the score on the test set.
''First`` evaluates the results using the predicted best segmentation for each image.
''First`` is only applicable to our method, since the segmentations from gPb-owt-ucm don't have associated scores to select a single segmentation. ODS is not applicable to our method, as \fgtiling \ generates independent segmentations. Note that ''First`` does not use any ground truth information to select the tiling to be evaluated for each image.

The BSDS dataset has multiple ground truth (human) segmentations for each image. To evaluate the quality of a segmentation, the average over all ground truth segmentations for that image is considered. As the provided human segmentations are different, the upper bound for OIS, ''First``, and ODS on the BSDS test set are 0.73. A score of 1.00 for BIS could be obtained by generating segments that perfectly cover all ground truth segments.

The results obtained by \fgtiling \  are competitive on BSDS and superior on the VOC2009. Note that the given VOC2009 scores are not using the ''segmentation challenge`` evaluation which requires recognition, but evaluating the quality of unlabeled segmentations like the method we compare with~\cite{contour_to_regions09}. The results of gPb-owt-ucm on VOC2009 have been computed by us using the code provided by the authors and are consistent with their published results on VOC2008.


\section{Conclusions}
\label{sec:concl}
We have proposed a mid-level computational learning and inference framework for image segmentation that tiles multiple figure-ground hypotheses into a complete interpretation.
The inference problem is formulated as searching for high-scoring maximal cliques in a graph connecting non-overlapping putative figure/ground hypotheses. Clique potentials are based on both intrinsic Gestalt segment quality and compatibilities among neighboring image segments, as derived from statistics of 3d scene boundaries. Learning is formulated as optimizing the ranking of the best-K hypotheses, directly on the testing error, measuring the overlap between image tilings and the ground truth human annotations.
We have empirically analyzed the performance of our learning and inference components and have shown that these achieve state of the art results in the Berkeley and the VOC2009 segmentation benchmarks. In the latter the proposed method improves on the state-of-the-art by 28\% when considering the full set of generated tilings, and by 16\% for the predicted best tiling.
In future work we plan to combine segmentation and partial recognition in order to be able to interpret images that contain both familiar and unknown objects.


\begin{figure}[tbp]
\label{last_results}
\centering
\mbox{
\epsfig{figure=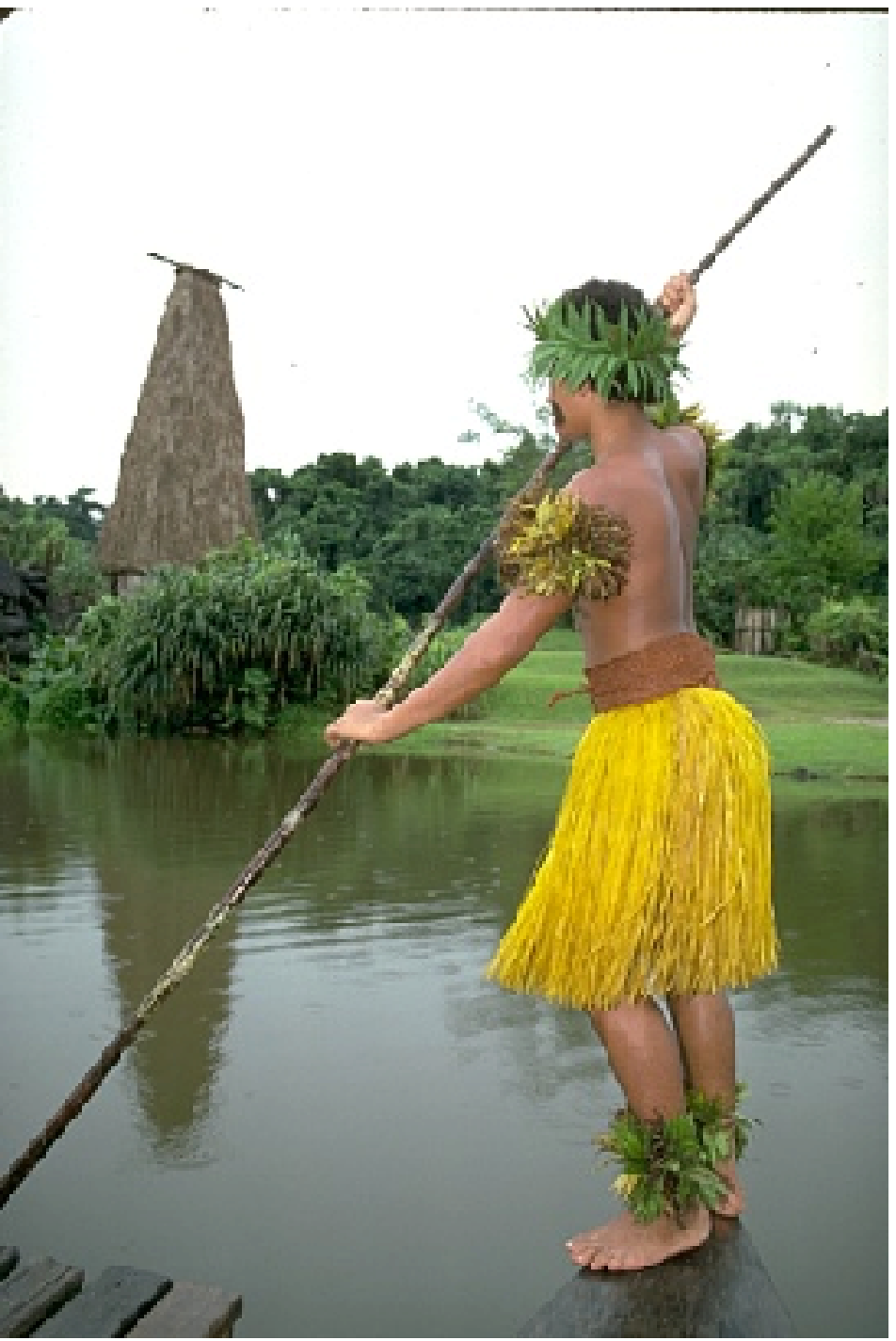,width=0.18\textwidth}~
\epsfig{figure=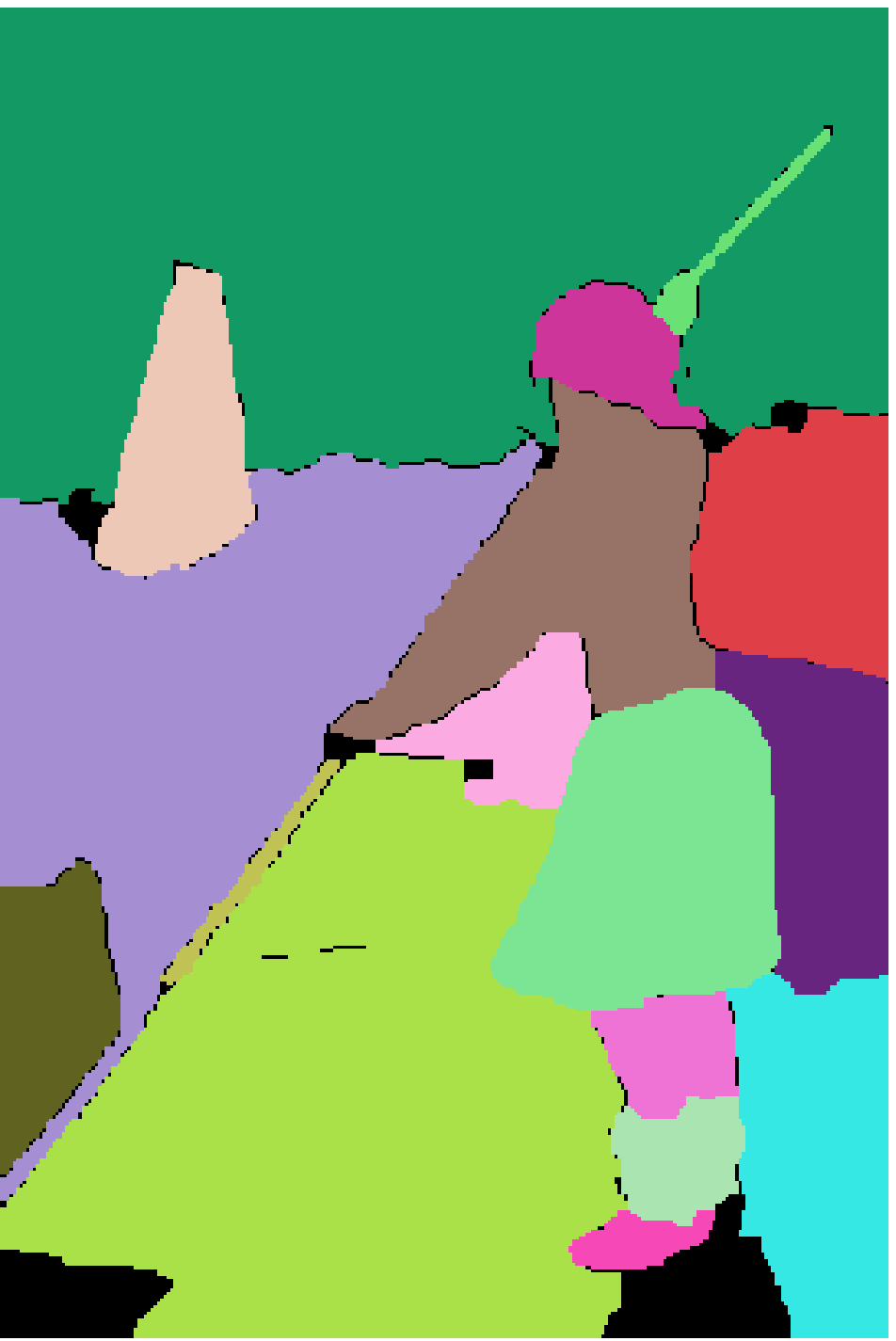,width=0.18\textwidth}~
\epsfig{figure=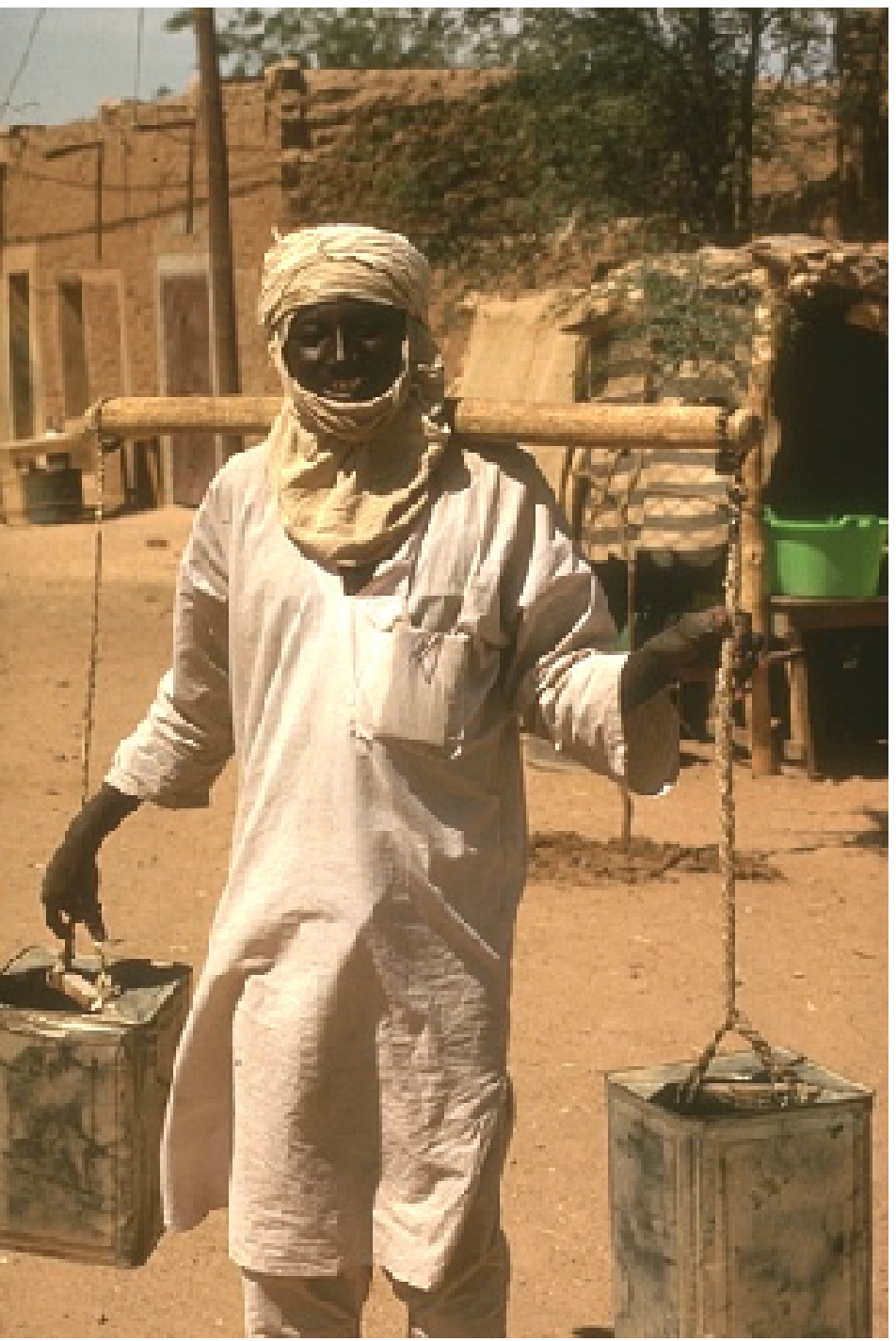,width=0.18\textwidth}~
\epsfig{figure=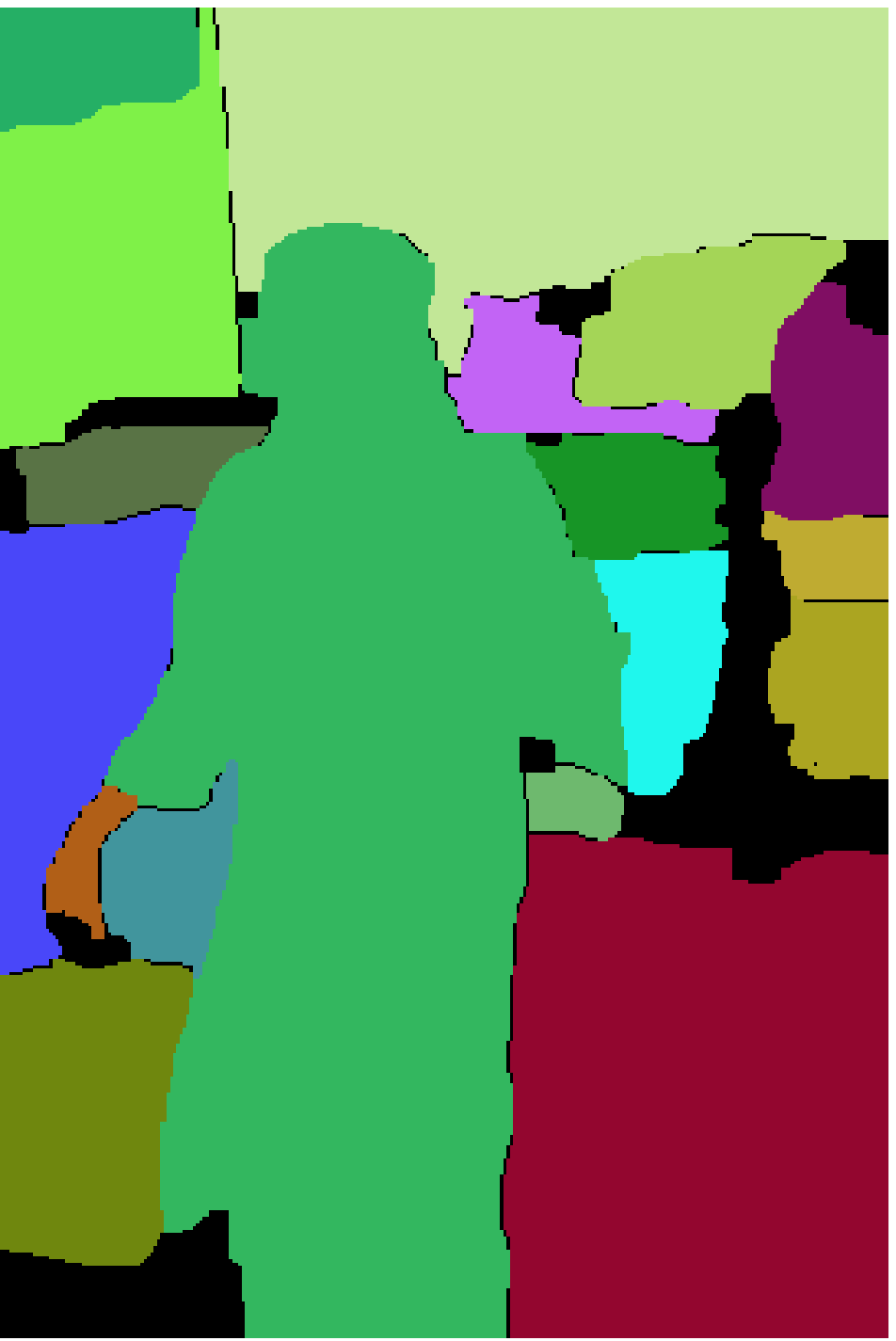,width=0.18\textwidth}~
}

\mbox{
\epsfig{figure=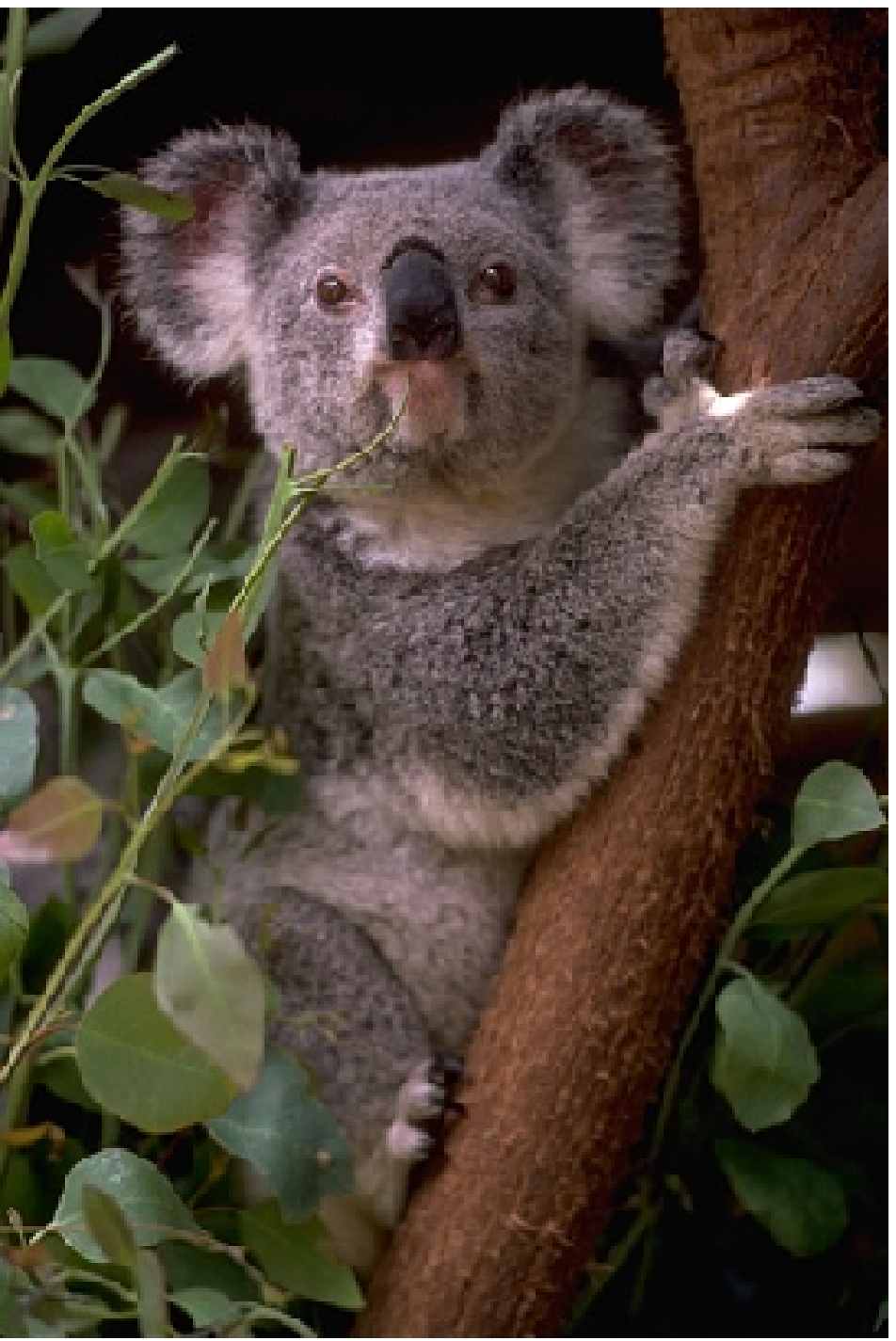,width=0.18\textwidth}~
\epsfig{figure=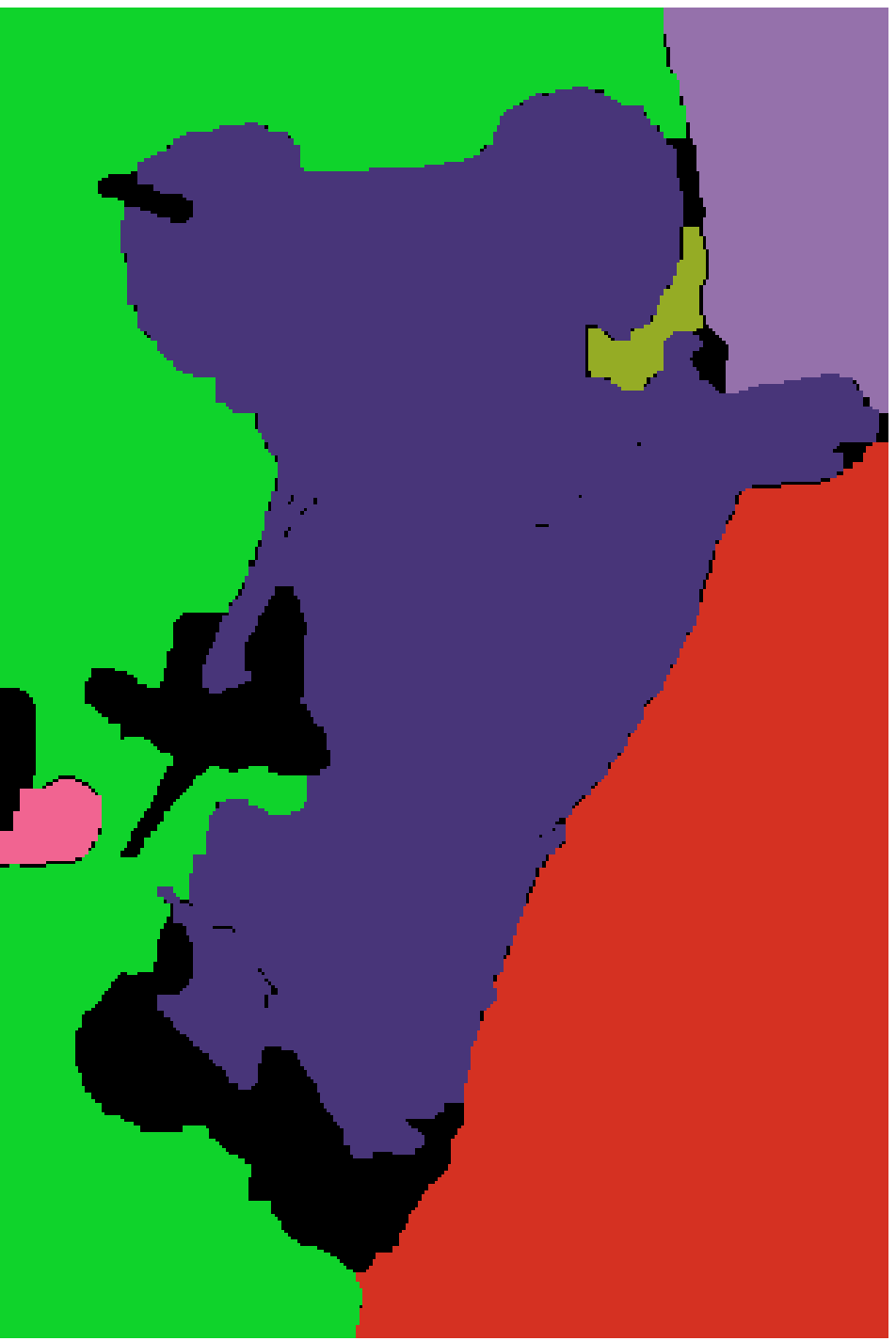,width=0.18\textwidth}~
\epsfig{figure=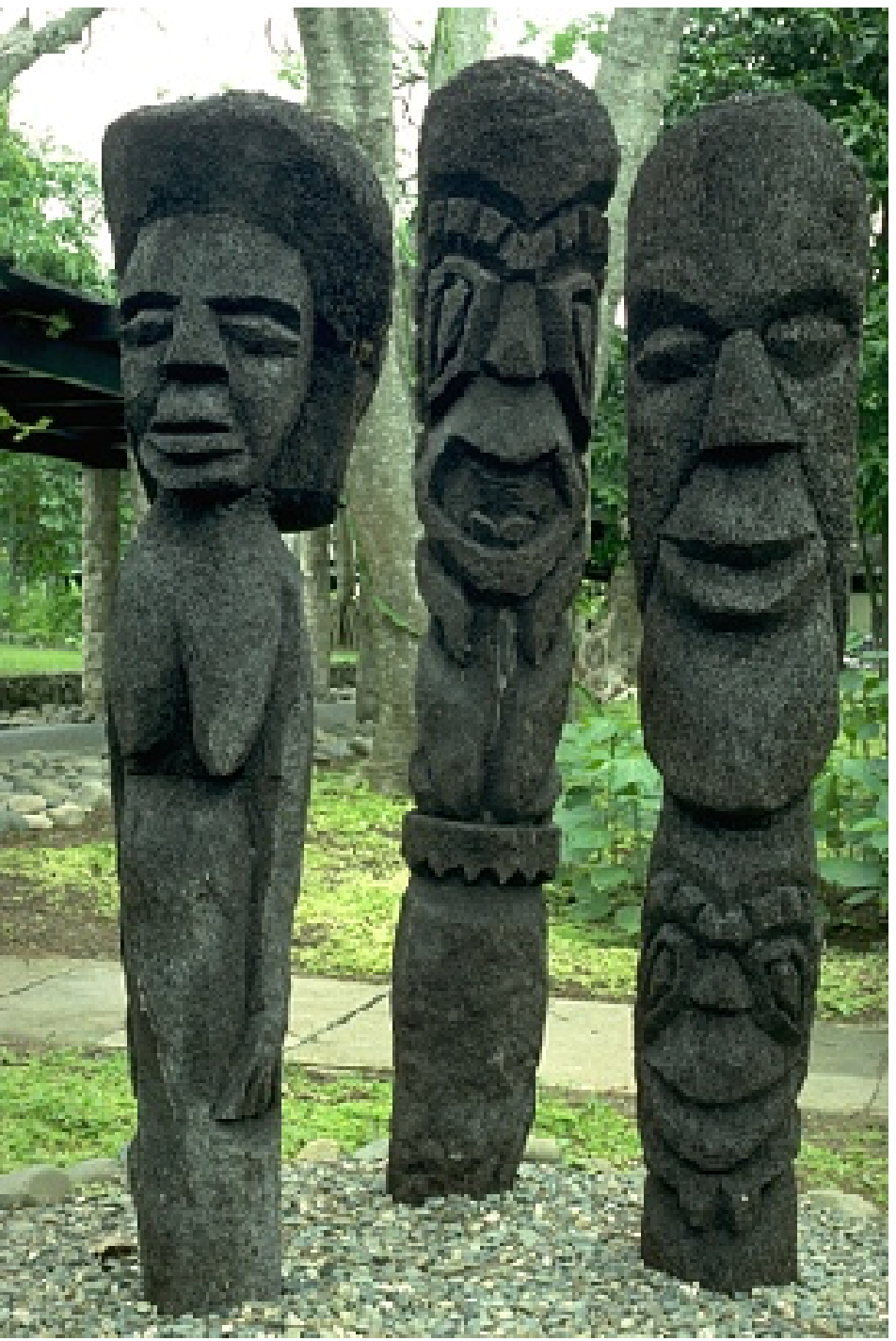,width=0.18\textwidth}~
\epsfig{figure=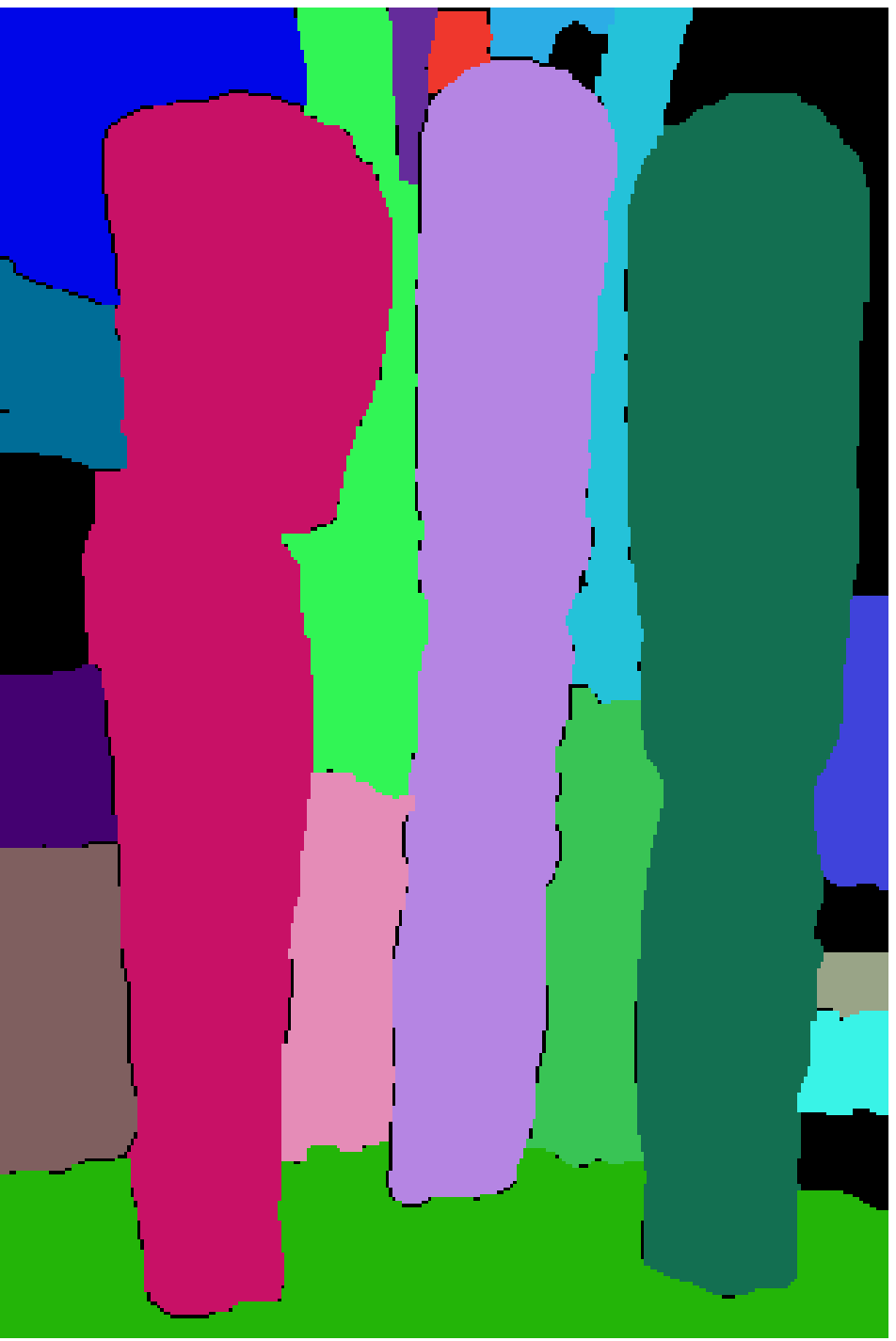,width=0.18\textwidth}~
}

\mbox{
\epsfig{figure=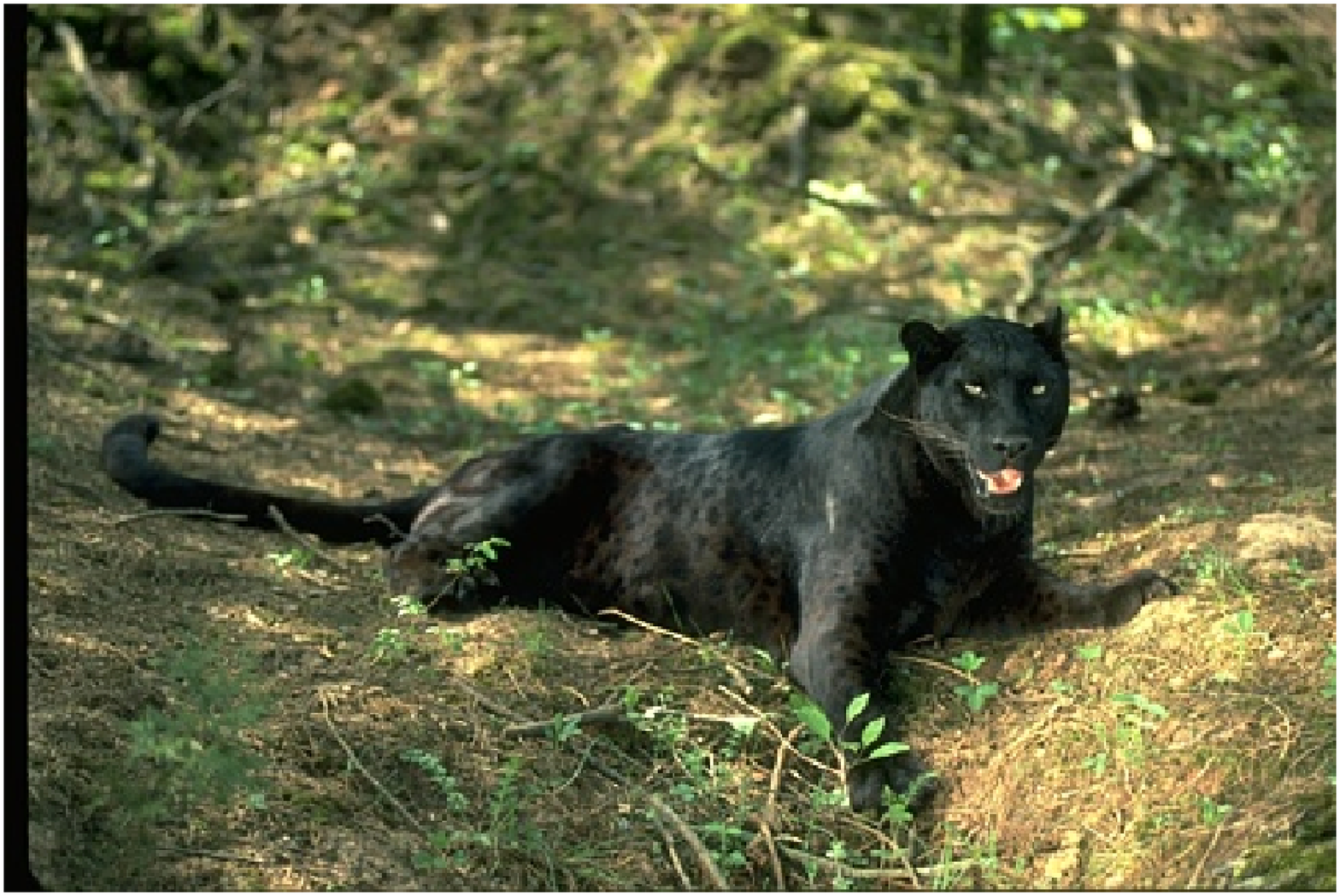,width=0.18\textwidth}~
\epsfig{figure=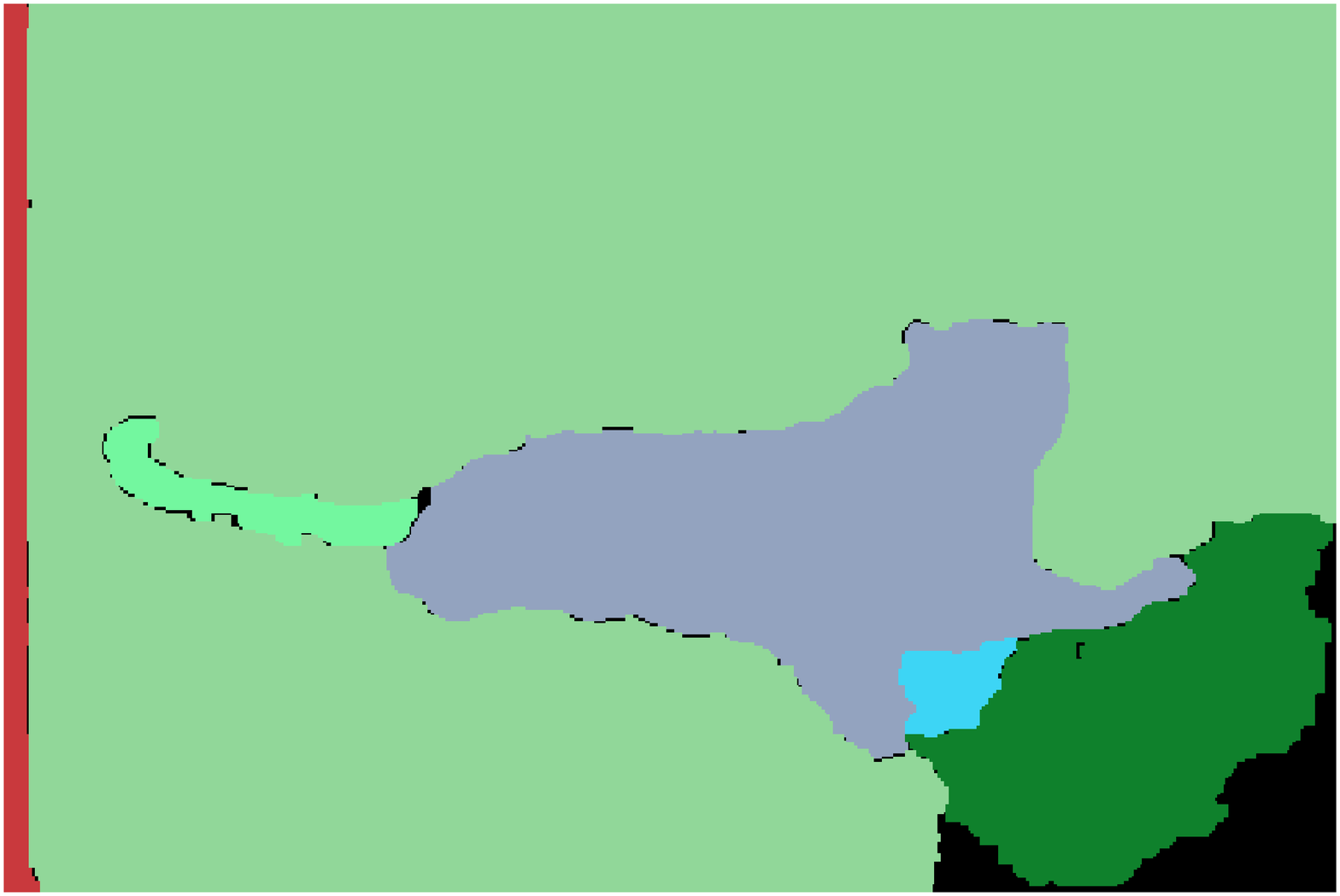,width=0.18\textwidth}~
\epsfig{figure=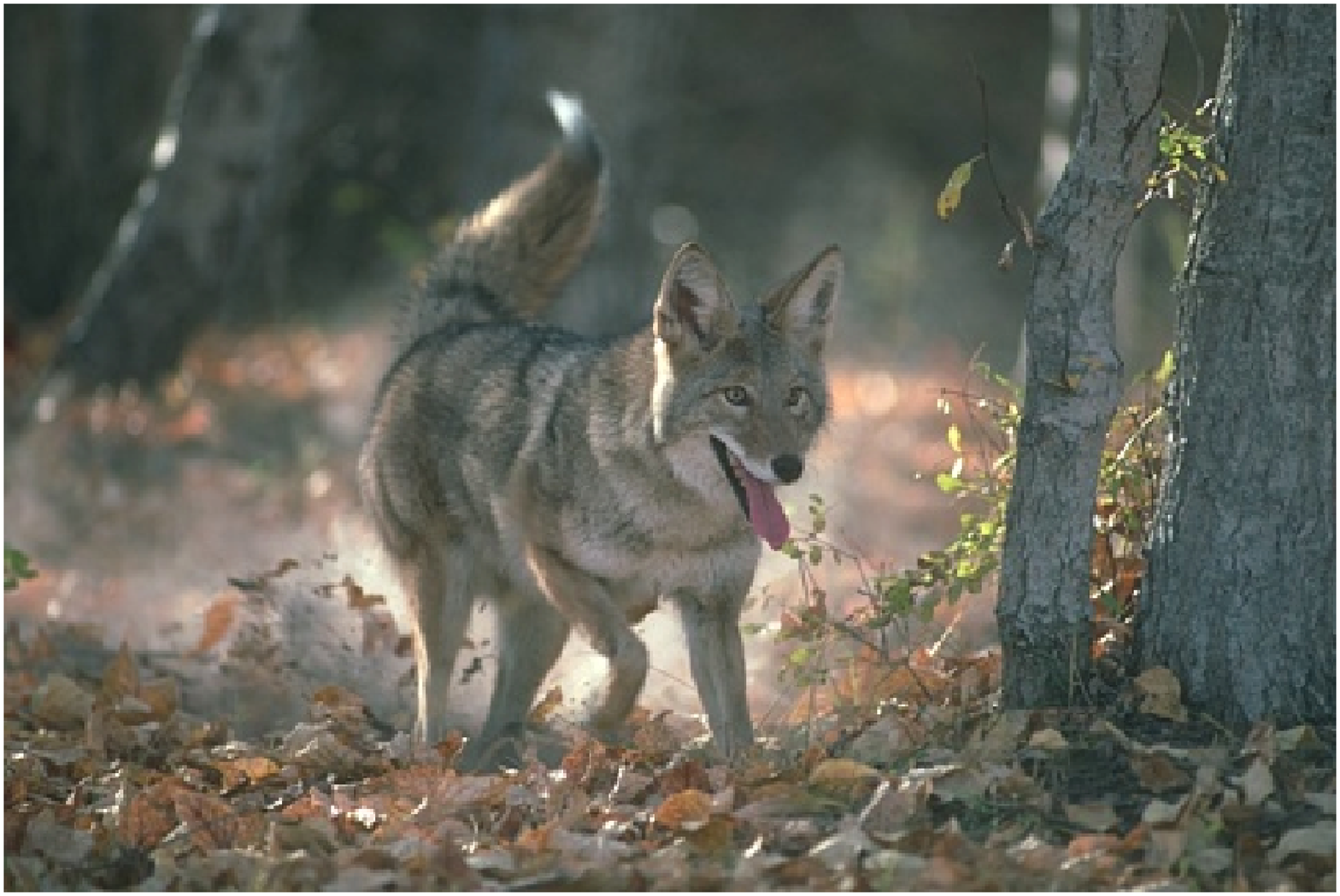,width=0.18\textwidth}~
\epsfig{figure=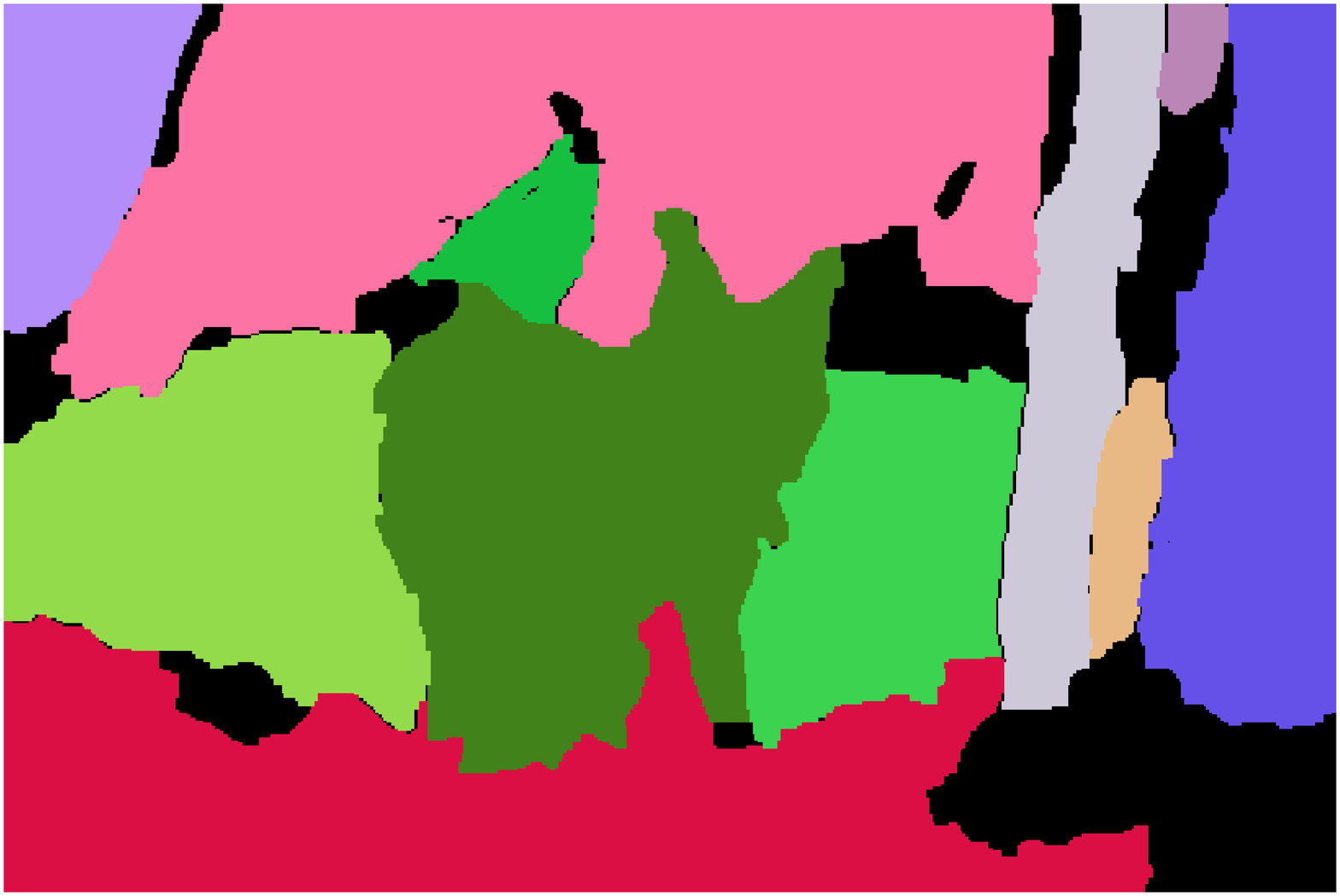,width=0.18\textwidth}~
}

\caption{{\it (Best viewed in color)} Images from the test set of BSDS, accompanied by our highest quality tiling. Areas of the image not covered are colored in black.}\label{fig:bsds_tilings}
\end{figure}

\begin{figure}[tbp]
\centering
\mbox{
\epsfig{figure=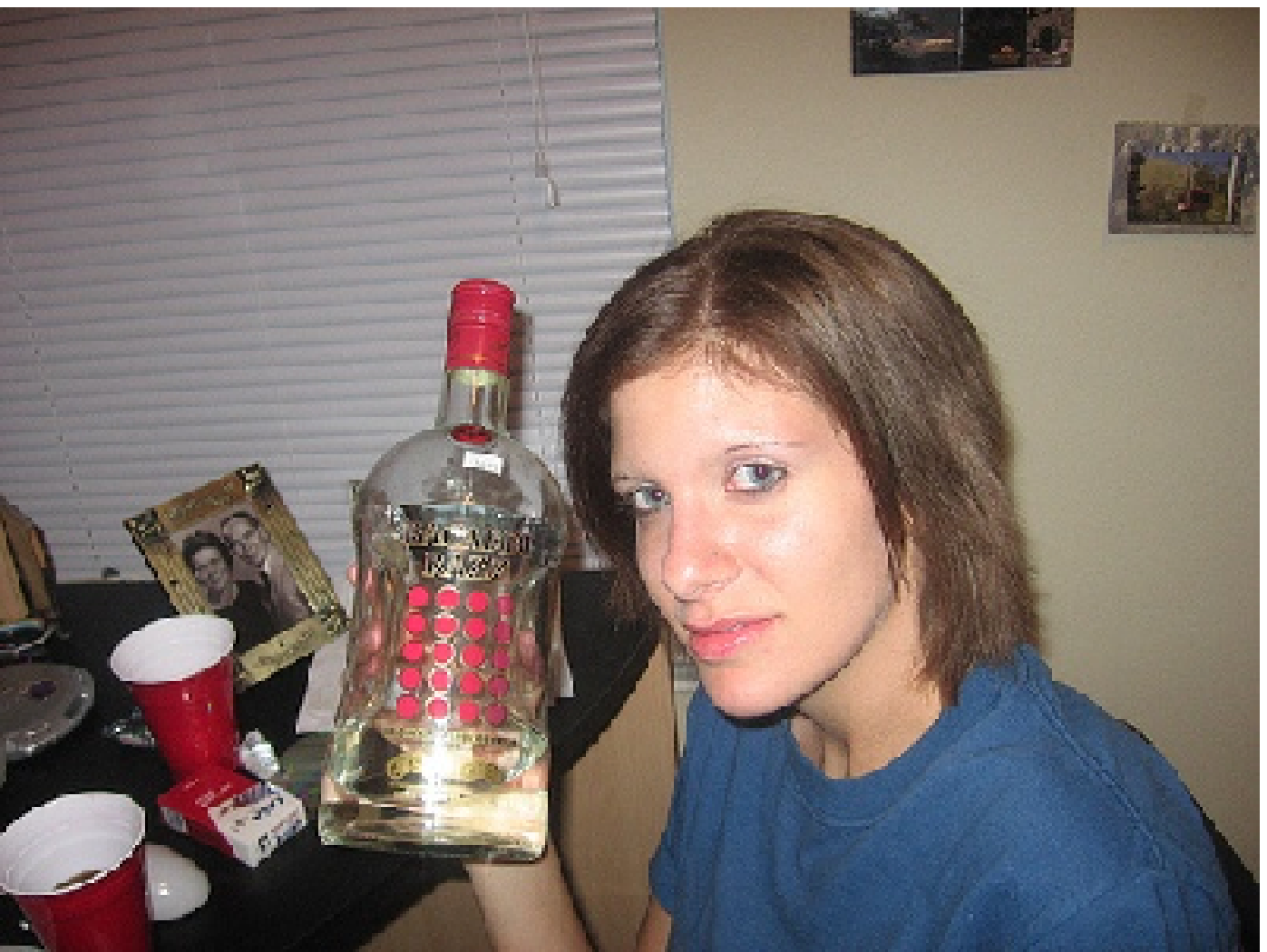,width=0.18\textwidth}~
\epsfig{figure=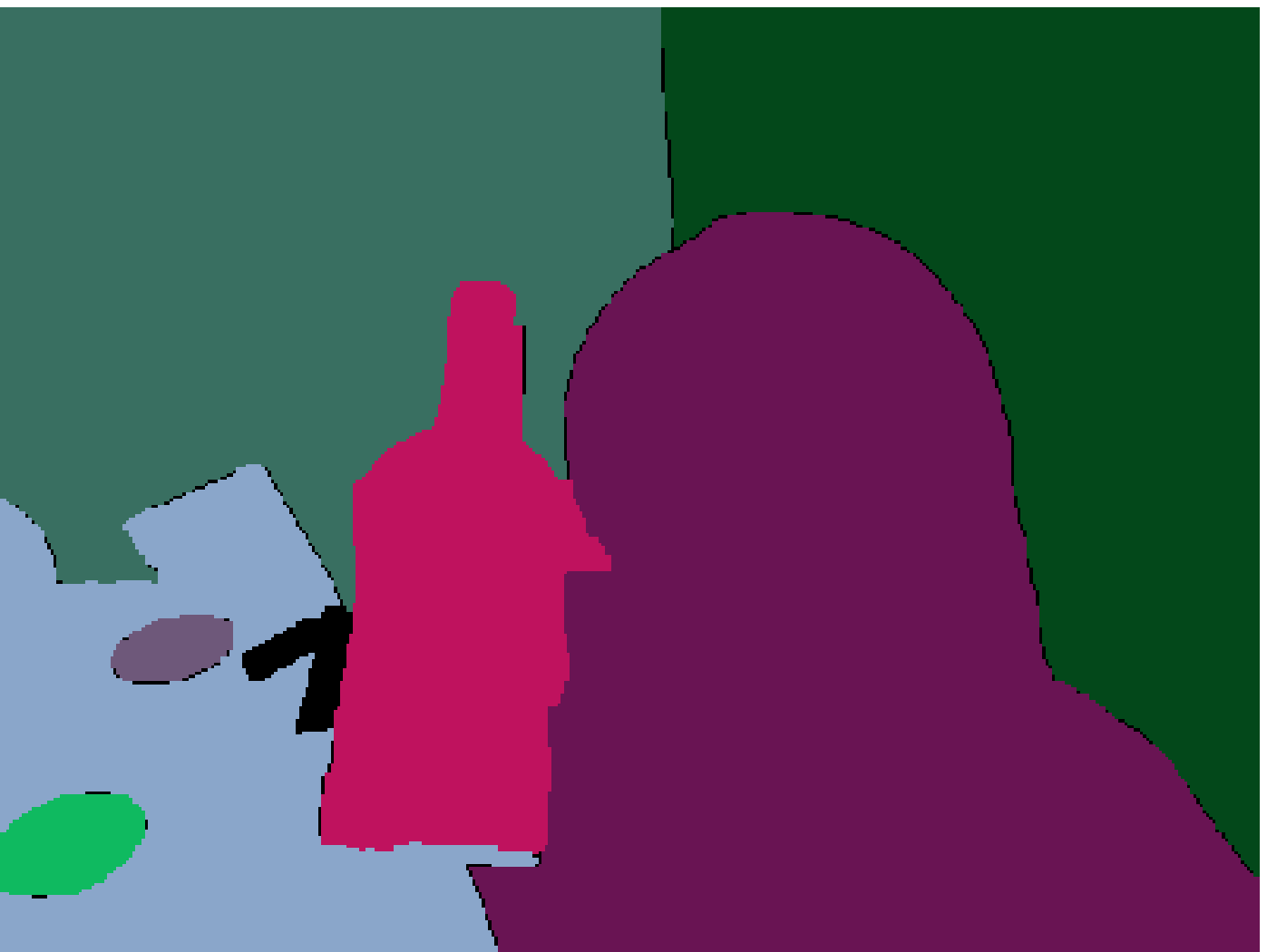,width=0.18\textwidth}~
\epsfig{figure=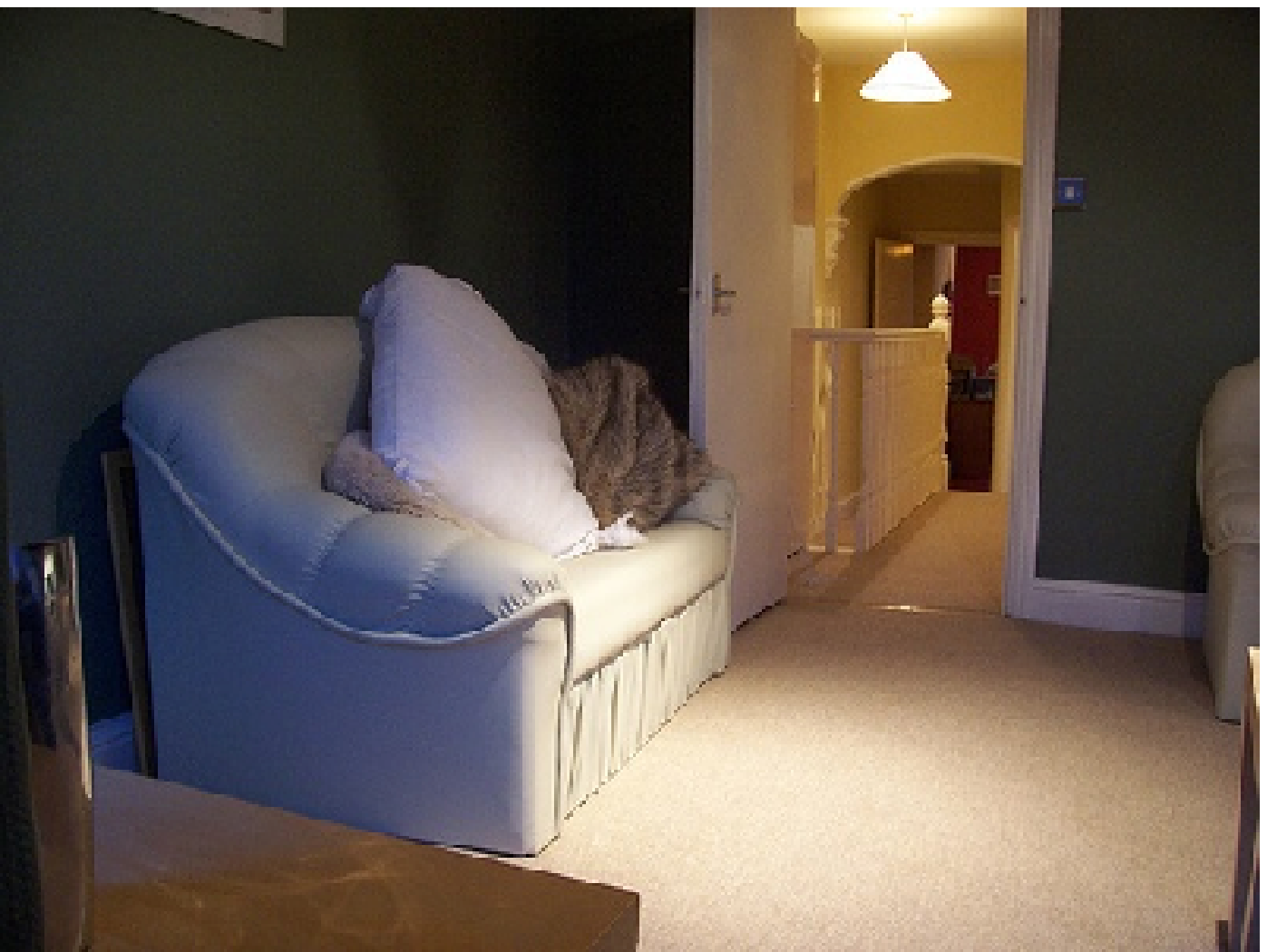,width=0.18\textwidth}~
\epsfig{figure=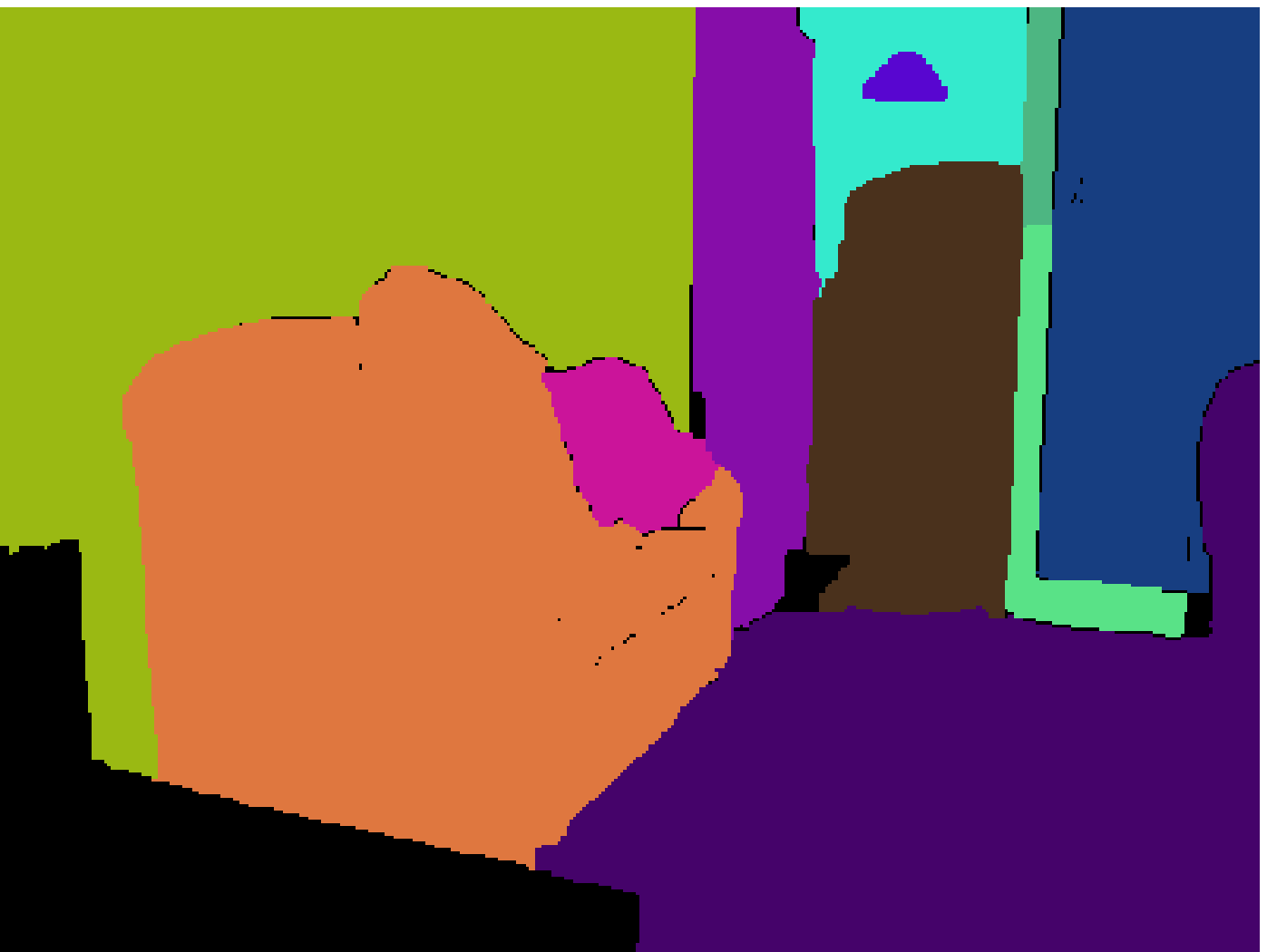,width=0.18\textwidth}~

}

\mbox{
\epsfig{figure=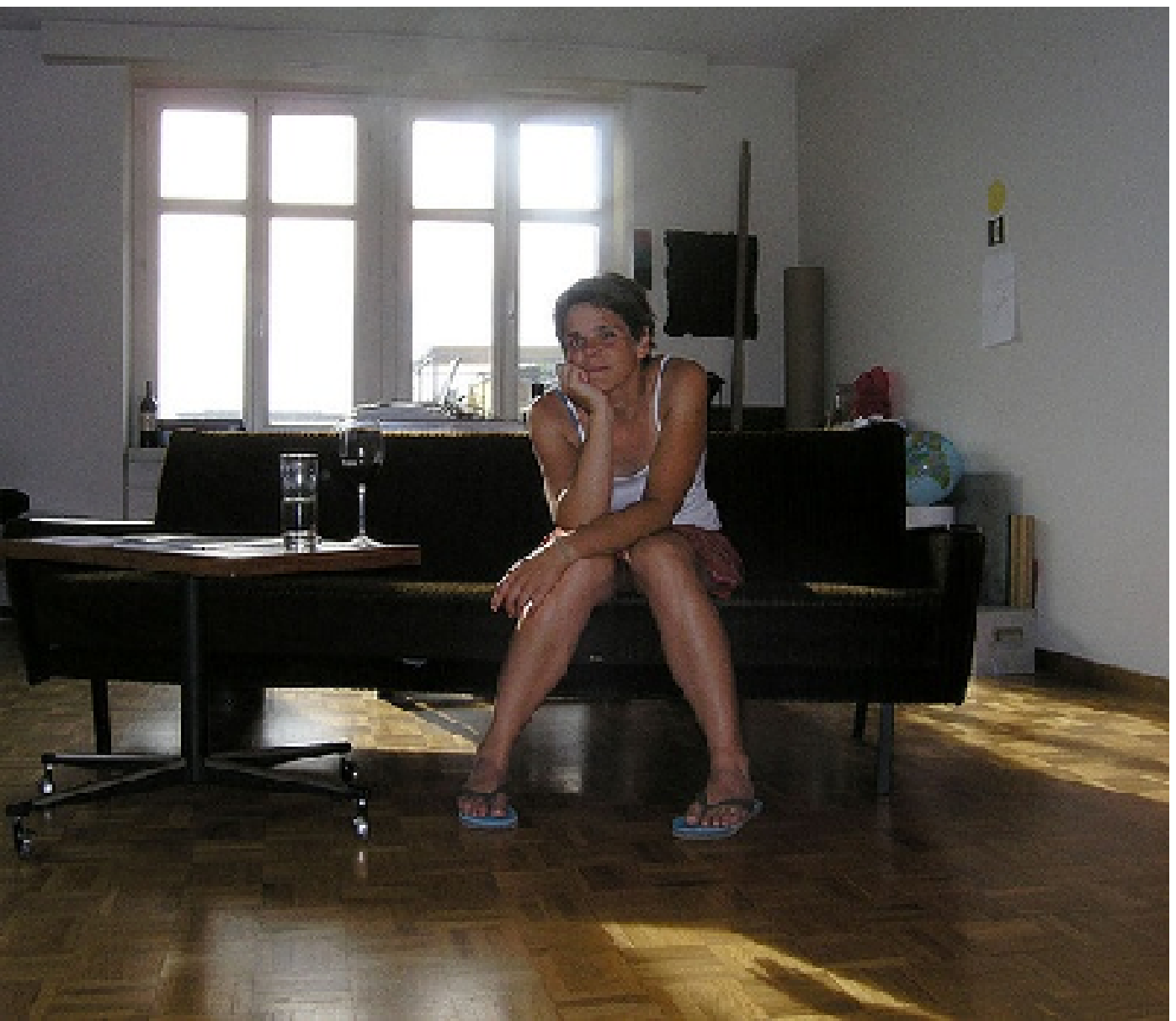,width=0.18\textwidth}~
\epsfig{figure=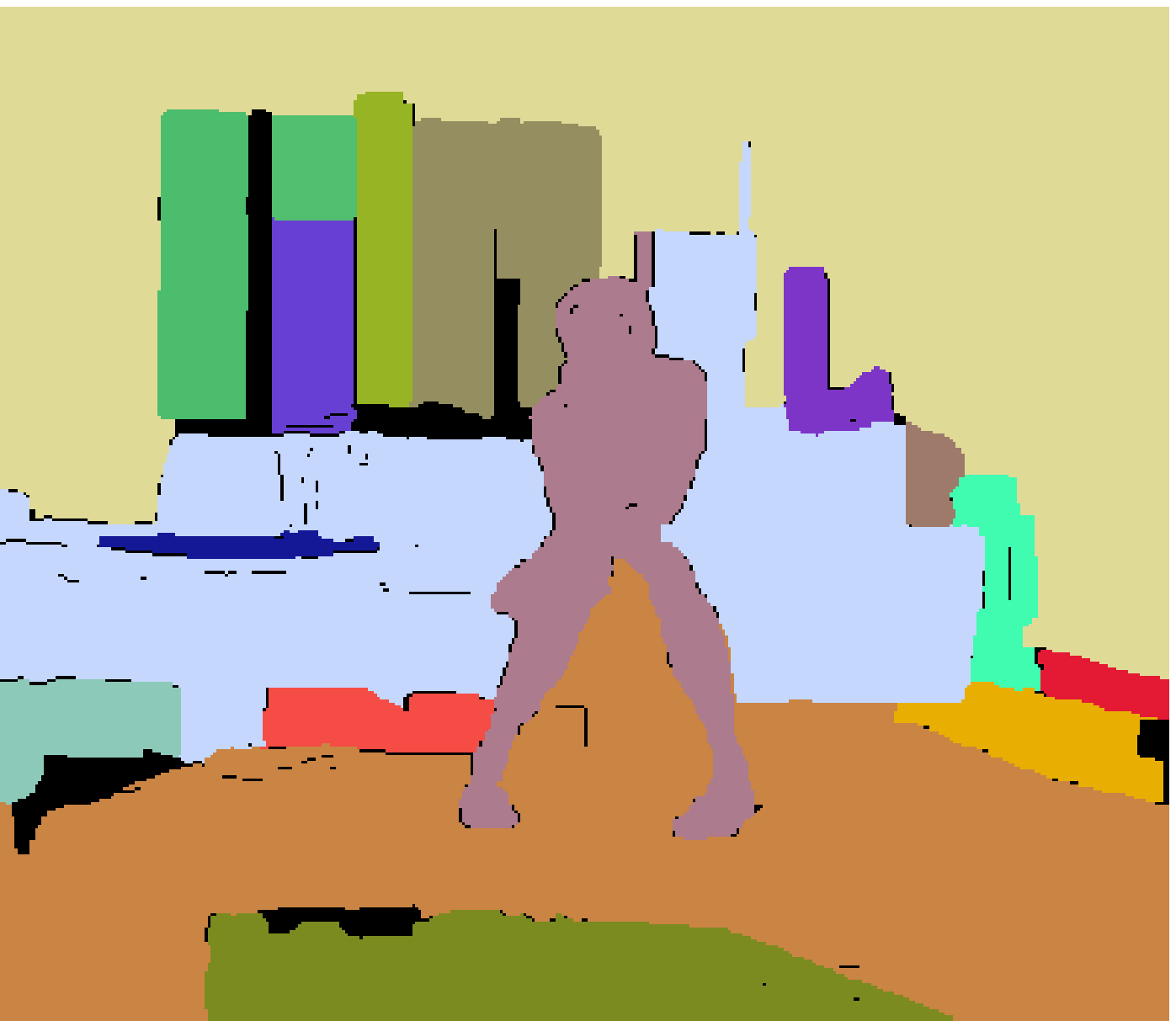,width=0.18\textwidth}~
\epsfig{figure=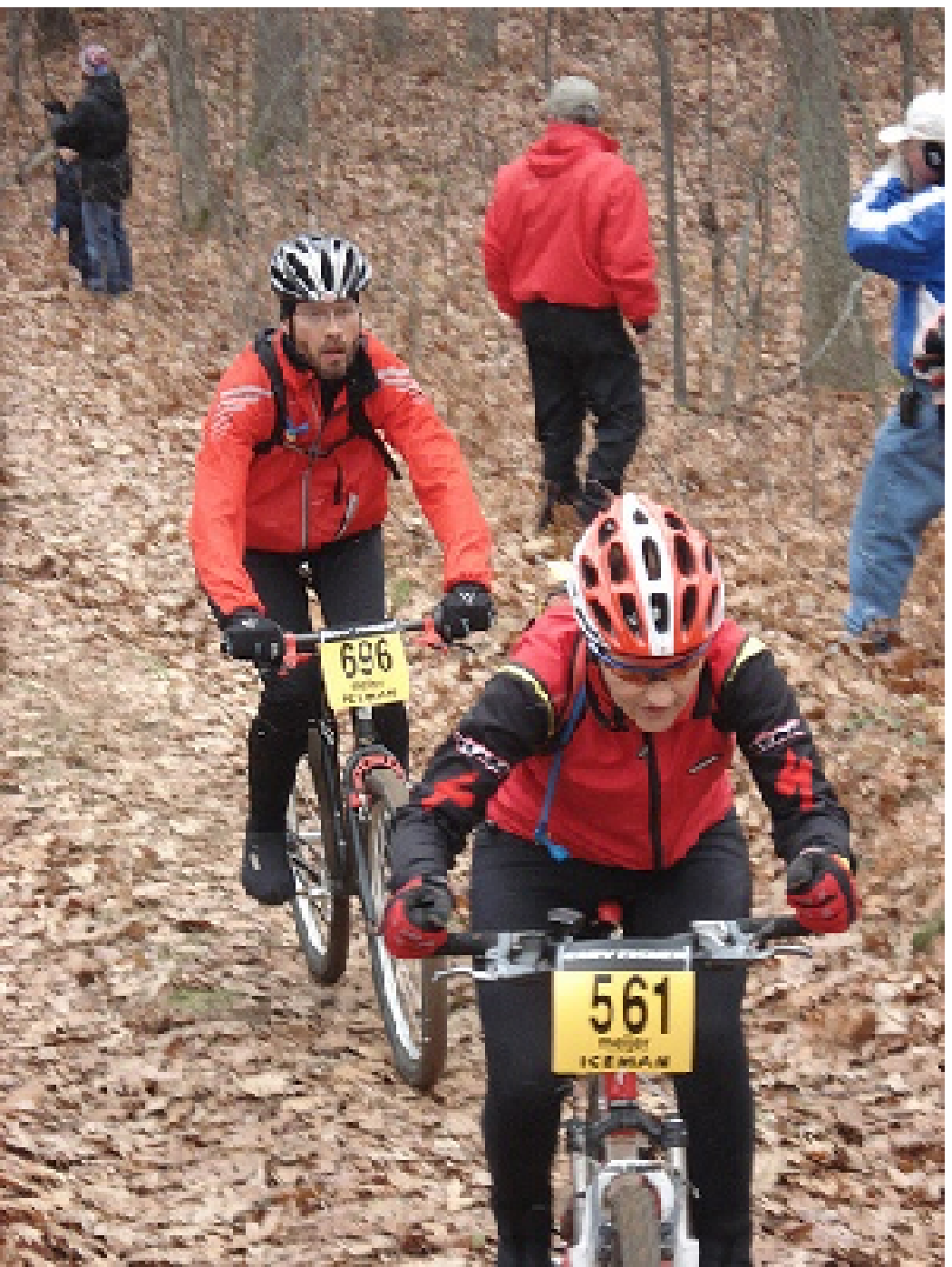,width=.18\textwidth}~
\epsfig{figure=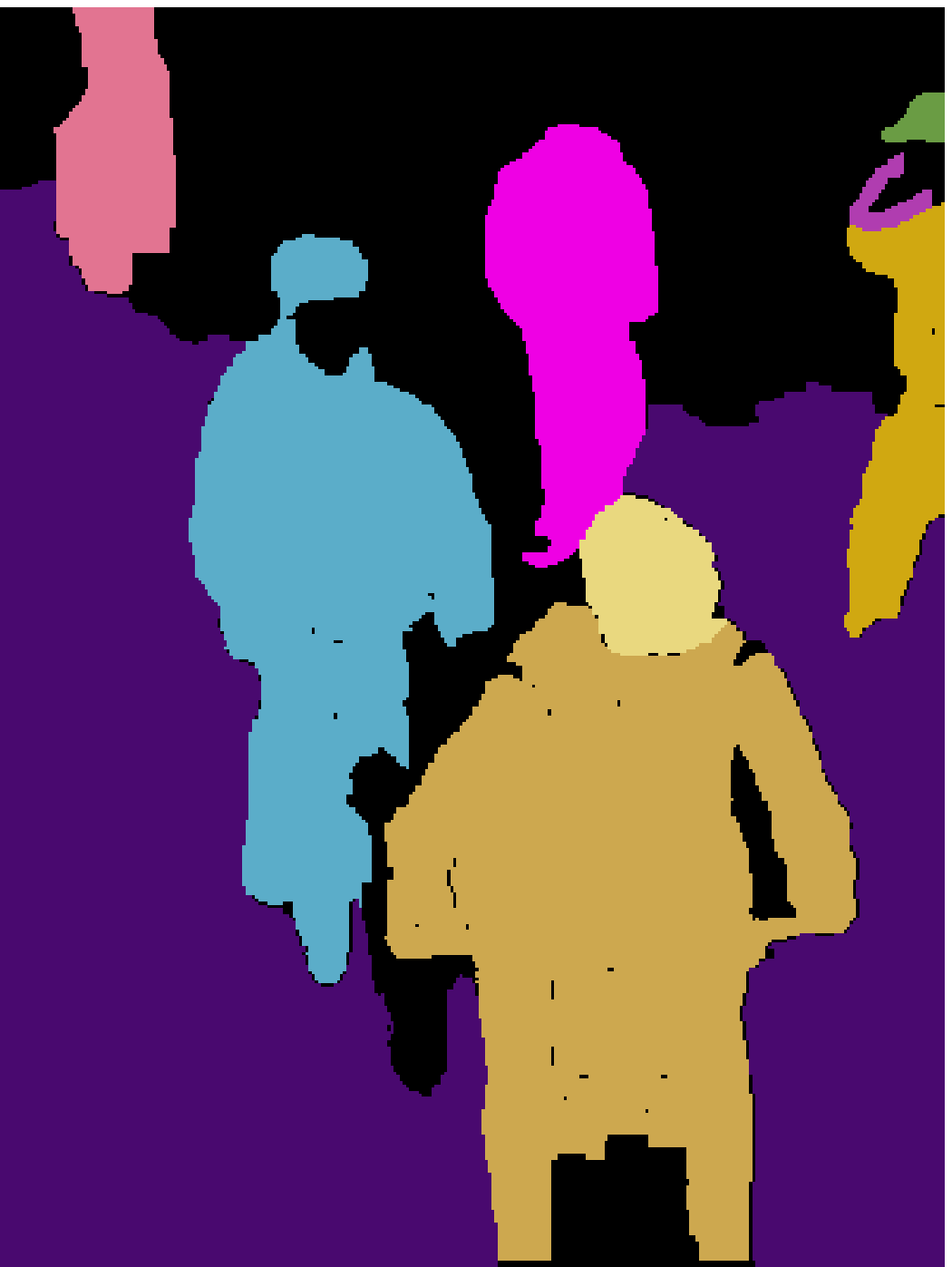,width=.18\textwidth}~
}

\mbox{
\epsfig{figure=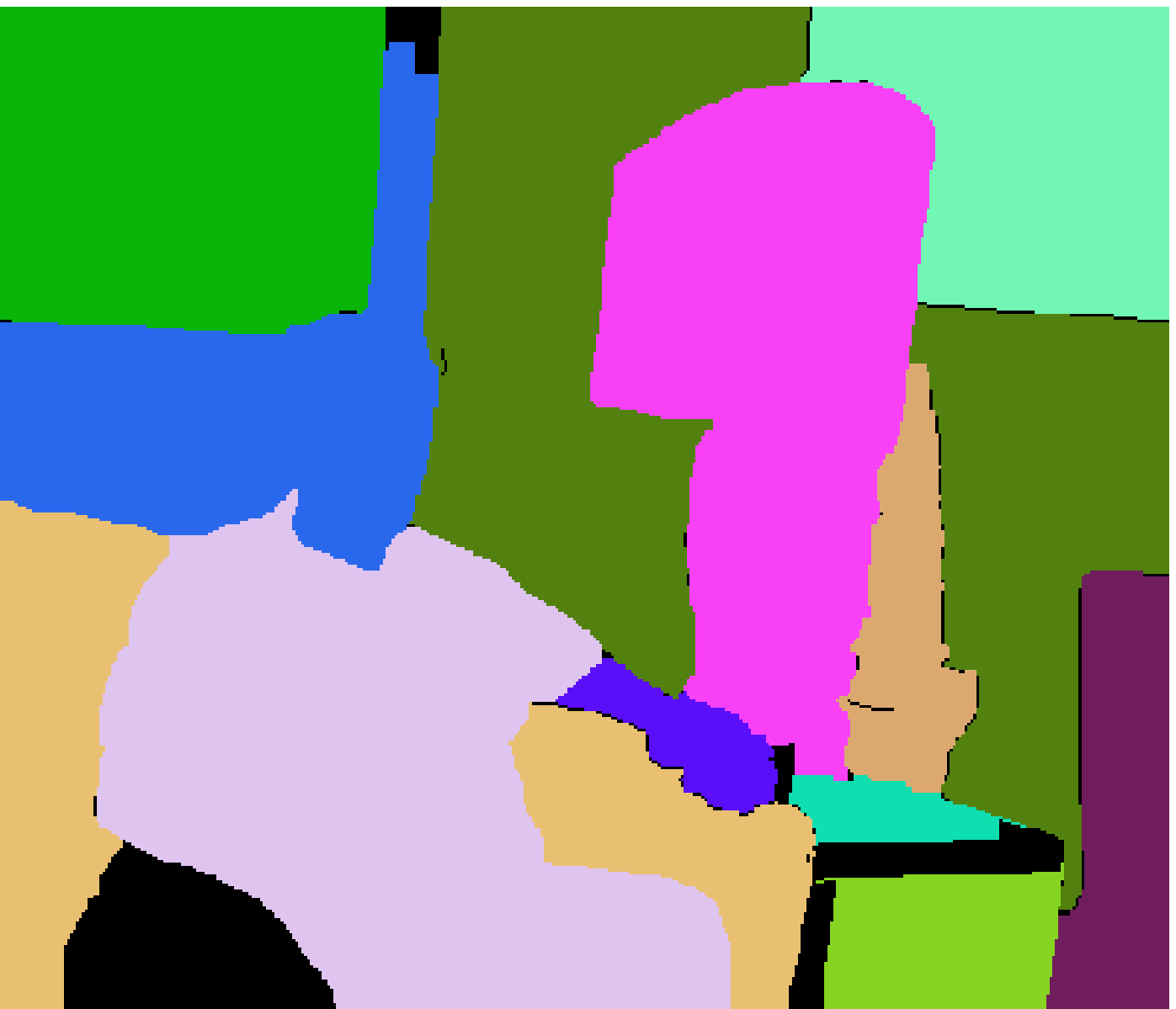,width=.18\textwidth}~
\epsfig{figure=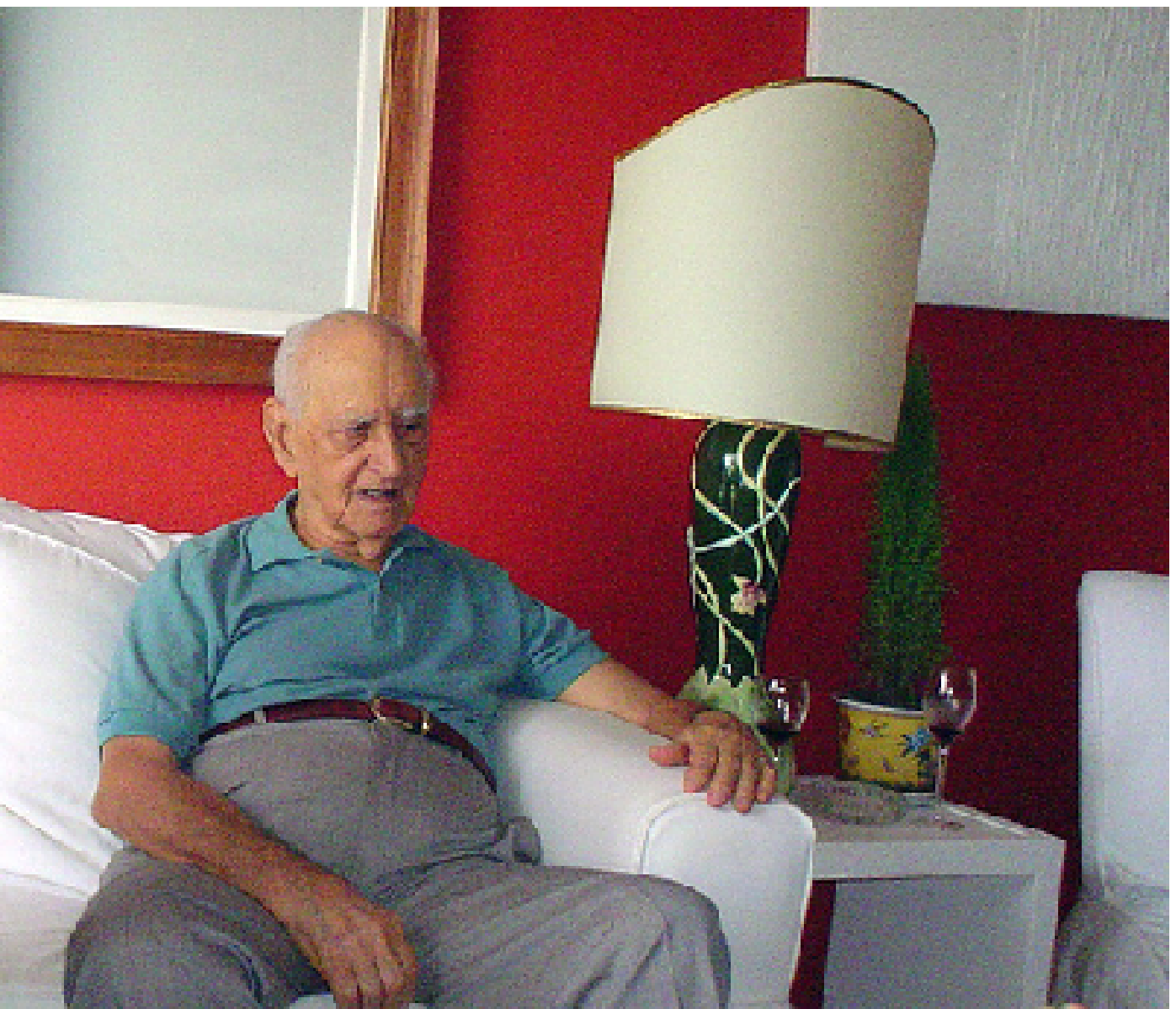,width=.18\textwidth}~
\epsfig{figure=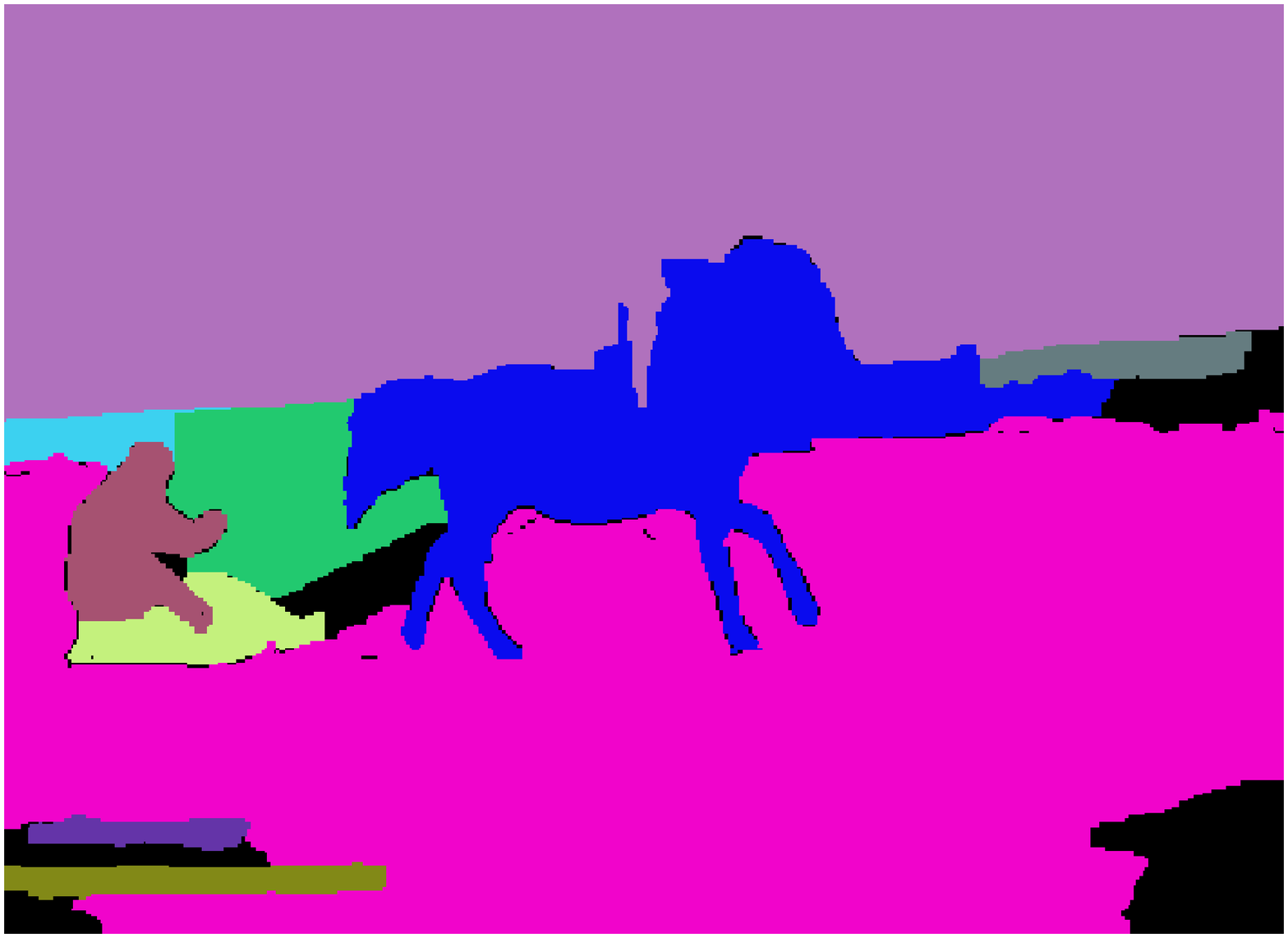,width=0.18\textwidth}~
\epsfig{figure=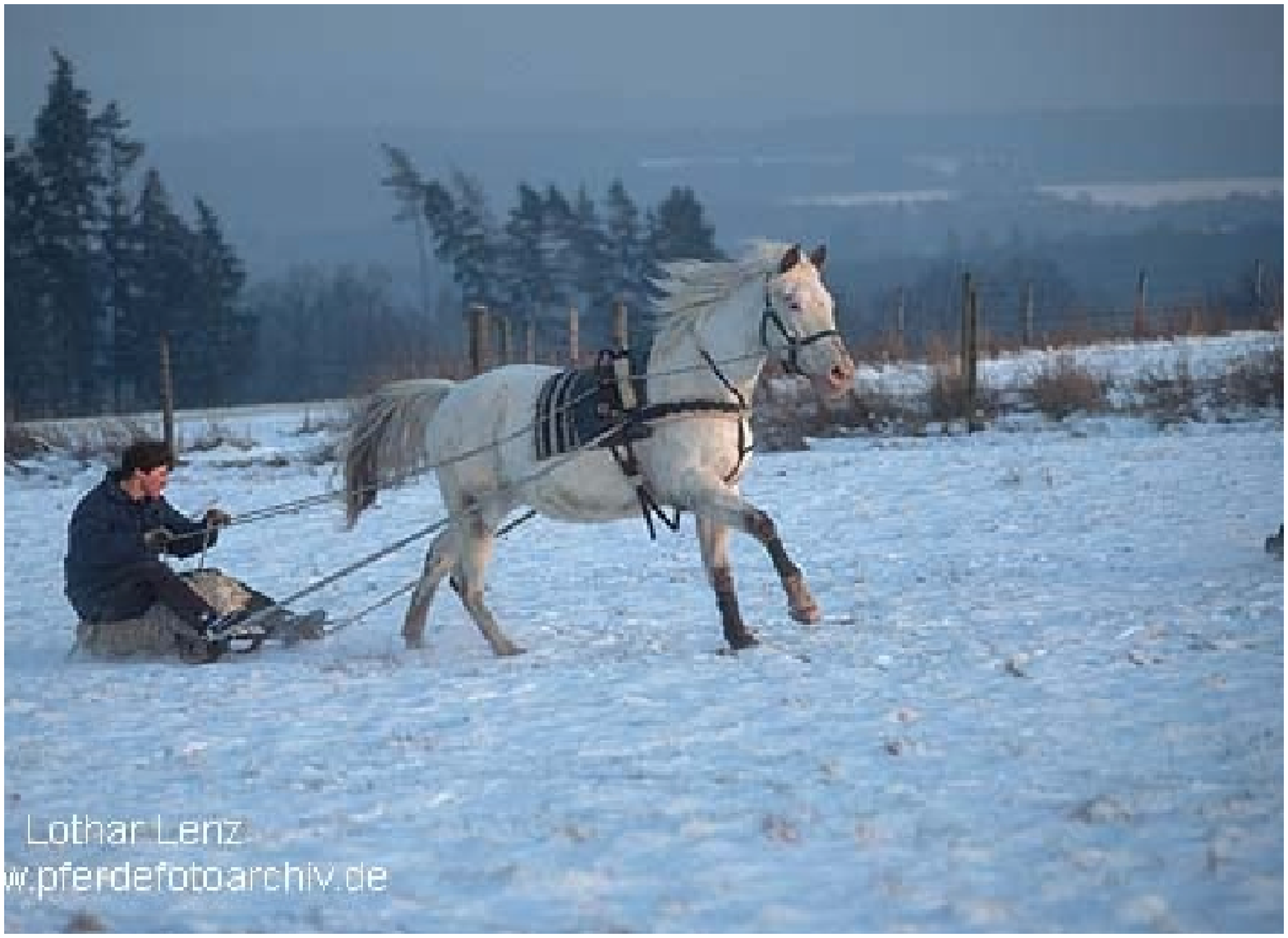,width=0.18\textwidth}~
}

\mbox{
\epsfig{figure=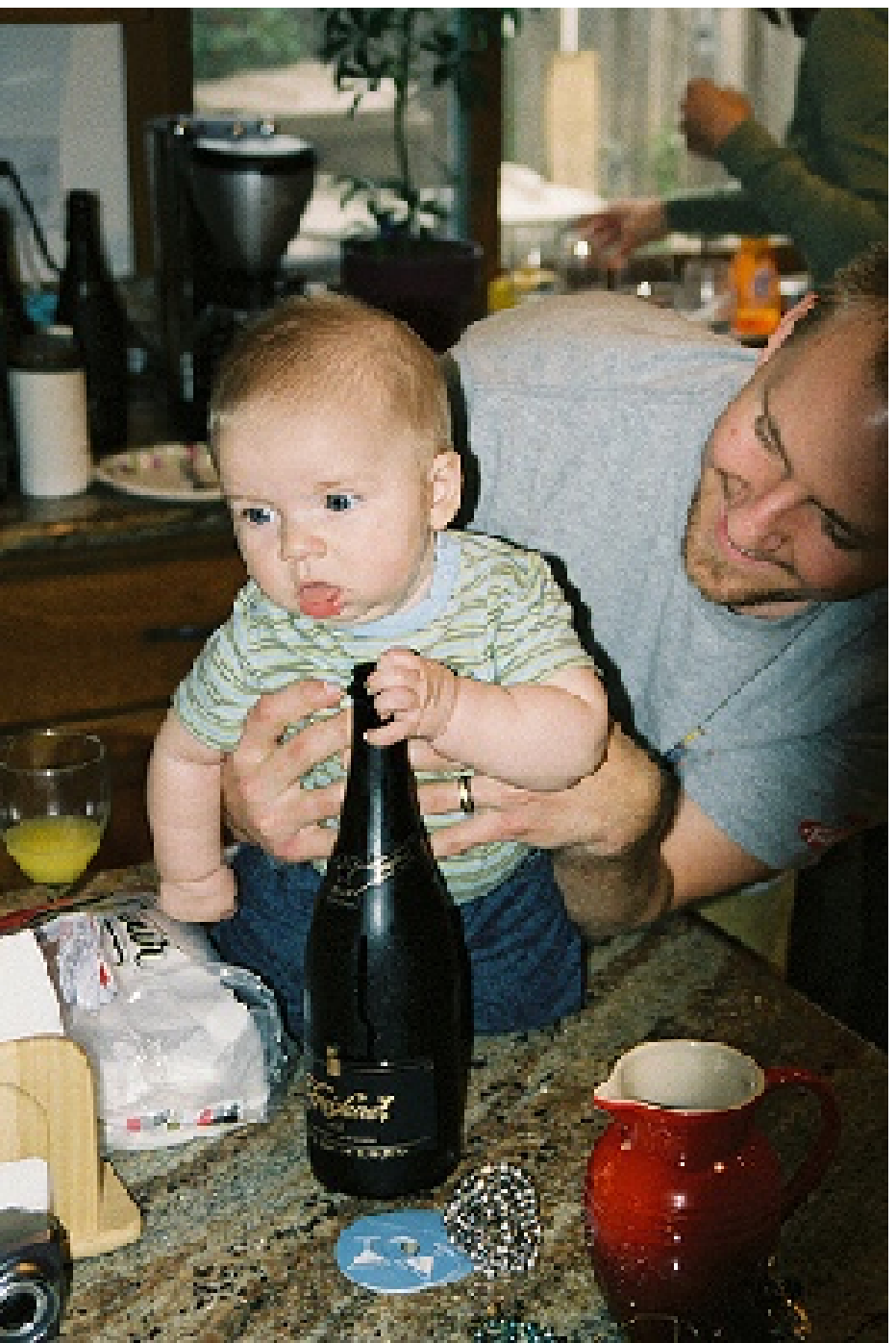,width=.18\textwidth}~
\epsfig{figure=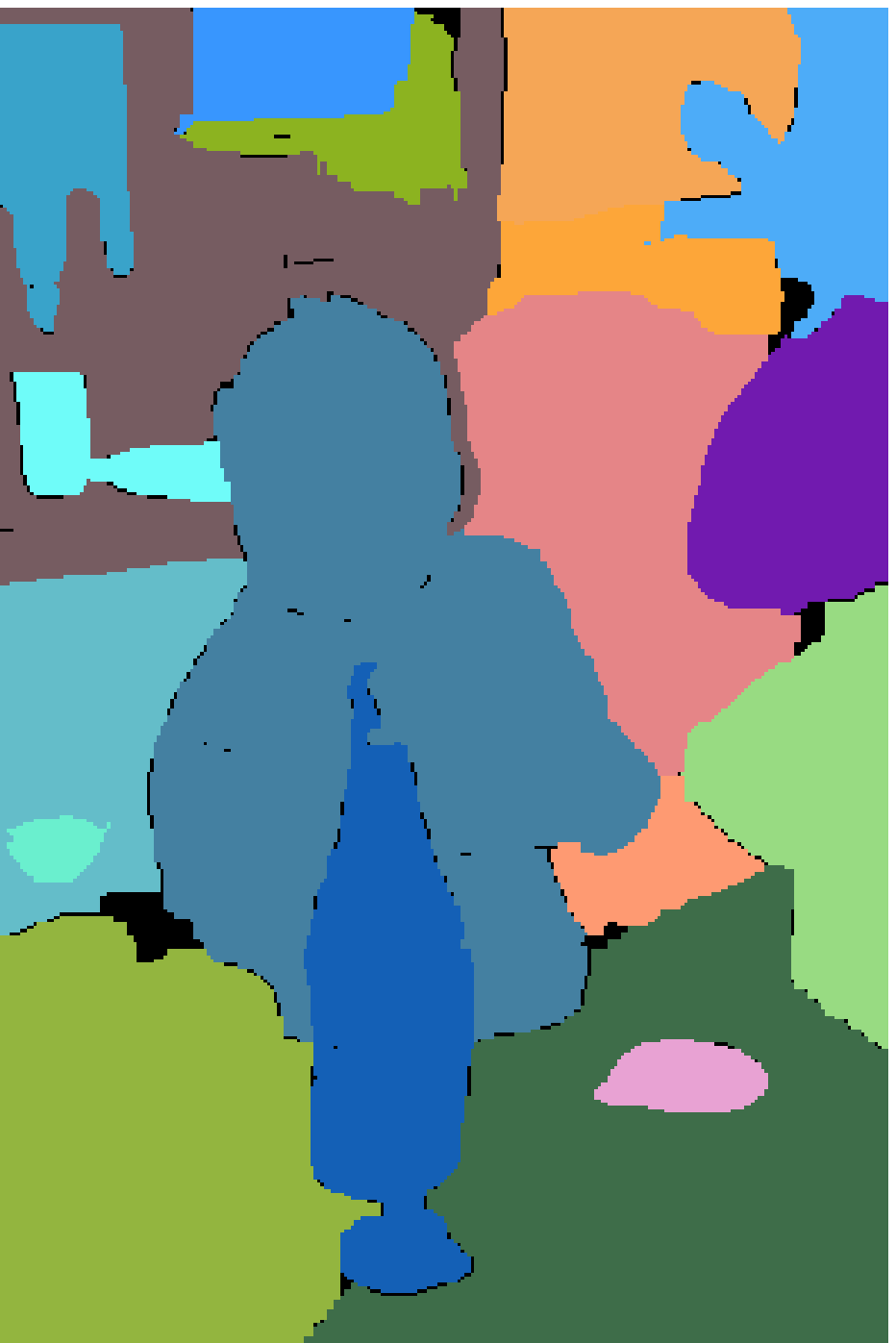,width=.18\textwidth}~
\epsfig{figure=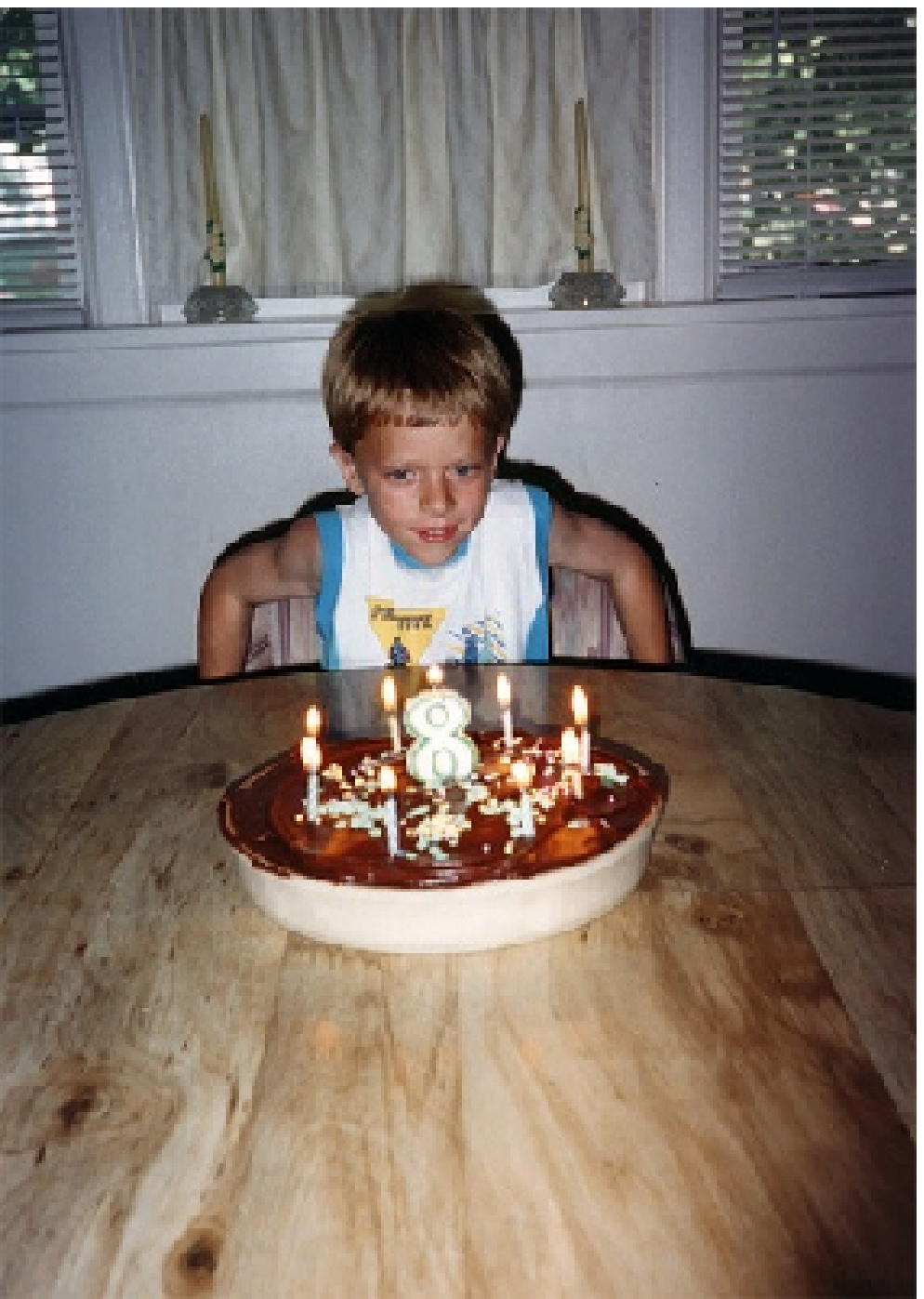,width=.18\textwidth}~
\epsfig{figure=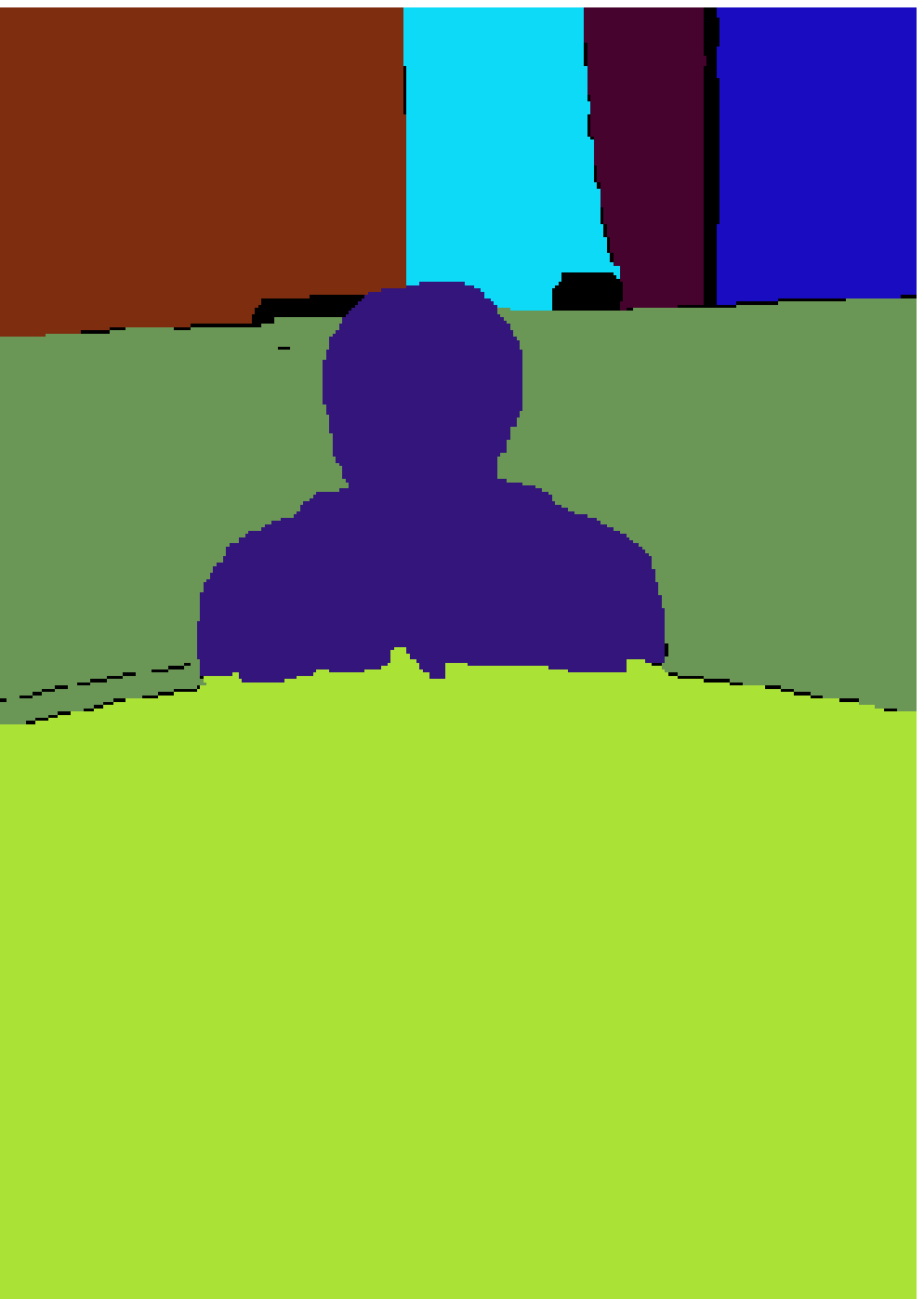,width=.18\textwidth}~
}
\label{fig:voc_images}
\caption{{\it (Best viewed in color)} Images from the validation set of VOC2009, accompanied by our highest quality segmentation.
Non covered areas of the image are colored in black. We colored disconnected segments with the same color, when they have strong occlusion cues with a same third segment, and well as similar color and texture.}
\label{fig:voc_tilings}
\end{figure}

\end{document}